\documentclass[11pt]{article}

\usepackage[]{acl}

\usepackage{times}
\usepackage{latexsym}
\usepackage{kotex}

\usepackage[T1]{fontenc}

\usepackage[utf8]{inputenc}

\usepackage{microtype}

\usepackage{inconsolata}

\usepackage{graphicx}
\usepackage{booktabs}
\usepackage{multirow}
\usepackage{adjustbox}
\usepackage{tabularx}
\usepackage{xcolor}
\usepackage{colortbl}
\usepackage{float}
\usepackage{subcaption}
\usepackage{hyperref}
\usepackage{longtable}
\usepackage{array}
\usepackage{tcolorbox}
\definecolor{pmspos}{HTML}{1F5FBF}
\definecolor{pmsneg}{HTML}{C0392B}

\definecolor{gainbg}{RGB}{226,242,228}
\definecolor{netbg}{RGB}{250,244,220}
\definecolor{passbg}{gray}{0.92}

\usepackage{amsmath}
\usepackage{amssymb}
\usepackage{bbm}
\usepackage{enumitem}
\usepackage{lipsum}
\usepackage{cleveref}

\newcommand{\method}{\textsc{MAPLE}}

\definecolor{posweak}{HTML}{D6E4F0}    
\definecolor{posmid}{HTML}{8AB8E0}     
\definecolor{posstrong}{HTML}{3D7AB8} 
\definecolor{negweak}{HTML}{F5D6D6}    
\definecolor{negmid}{HTML}{E08A8A}    
\definecolor{negstrong}{HTML}{B83D3D}  

\newcolumntype{C}[1]{>{\centering\arraybackslash}p{#1}}
\newlength{\bucketw}
\newlength{\cellw}
\title{MAPLE: A Multi-Axis Parametric Knowledge Extraction Lens for LLM Evaluation}

\author{
    Minsuh Joo$^{1}$, Youna Kim$^{2}$, Choonghyun Park$^{2}$,\\ \textbf{Soo Kyung Kim$^{13}$, Hyunsoo Cho$^{13*}$}
\\
$^1$Department of AI, Ewha Womans University \quad $^2$ Seoul National University
\\$^3$Institute for Multiscale Matter and Systems, Ewha Womans University
\\
\texttt{\{judyjoo21, chohyunsoo\}@ewha.ac.kr}
}
  
\begin{document}

\maketitle

\renewcommand{\thefootnote}{\fnsymbol{footnote}}
\footnotetext[1]{\noindent Corresponding author}


\begin{abstract}
    LLMs absorb vast knowledge from training data, yet their inherently probabilistic nature offers no guarantee that this knowledge will surface in their outputs.
    To bridge this gap, Knowledge Extraction Enhancement (KEE) techniques such as activation steering, decoding adjustments, and prompting modify inference, potentially reinforcing existing predictions or surfacing less accessible answers.
    Yet their evaluations centered on aggregate accuracy, which obscures diverse effects by collapsing recovered failures and disrupted successes into a single score.
    We introduce \method, a unified framework for characterizing how KEE techniques affect parametric knowledge extraction.
    Its diagnostic metrics disentangle these effects at the surface, probability, and beyond-sampling levels by measuring answer recovery and disruption, gold-token rank shifts on unrecovered failures, and recovery of answers missed by fixed-budget sampling.
    We evaluate six KEE techniques spanning decoding, activation steering, and prompting across three model families: OLMo2, Qwen2.5, and Llama3.
    Our analysis reveals distinct effects and trade-offs across techniques, highlighting the value of \method\ for understanding and comparing KEE beyond aggregate accuracy.    
\end{abstract}

\section{Introduction}
\label{sec:sec1}

    \begin{figure}[t]
  \includegraphics[width=0.75\columnwidth]{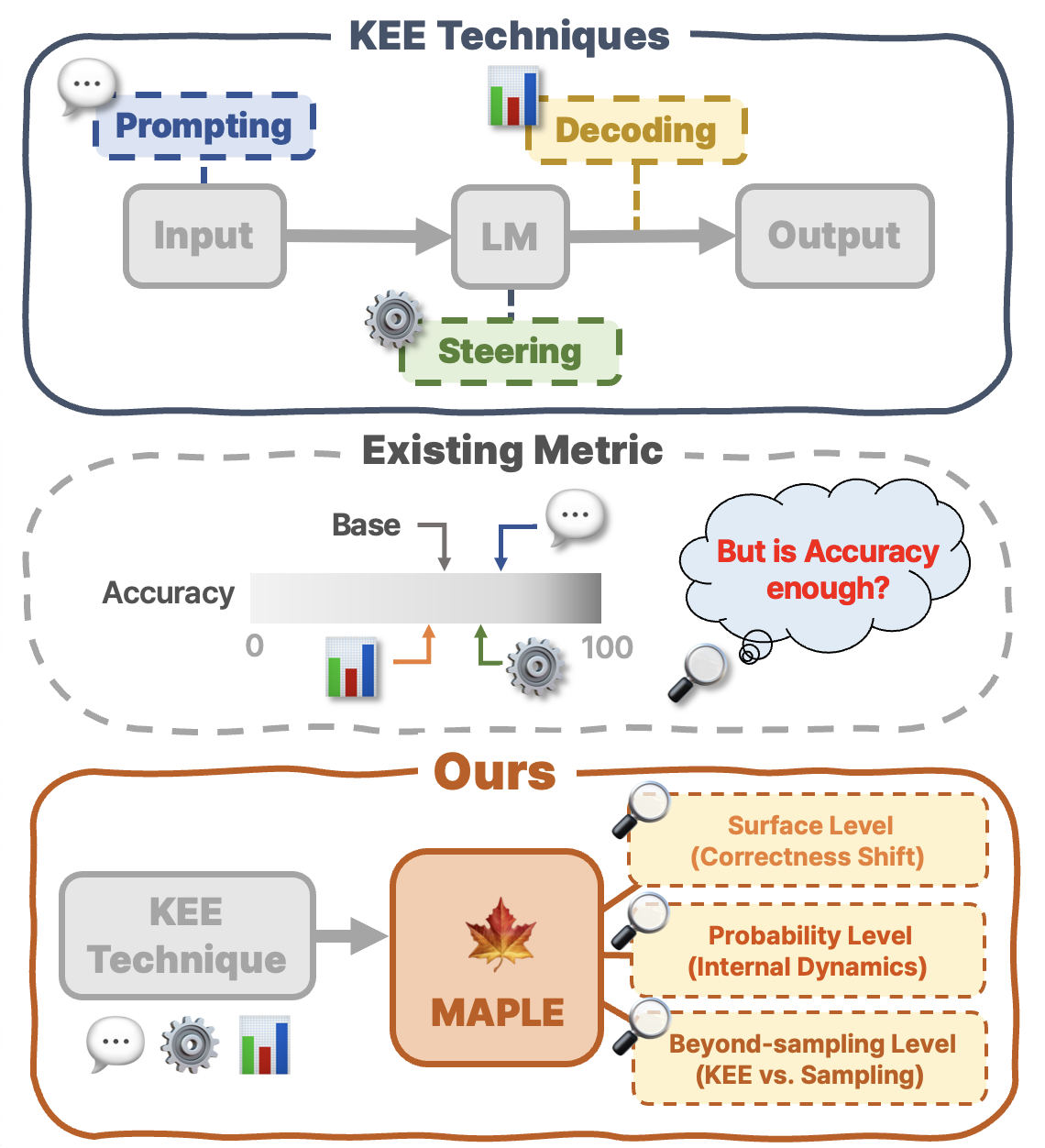} 
  \centering
  \caption {The overview of \method.}
  \label{fig:intro}
\end{figure}

    LLMs serve as vast repositories of factual knowledge, yet their outputs are not a direct readout of what they have learned~\cite{petroni2019language, jiang2020can}.
    Unlike traditional knowledge bases that return explicitly stored entries, LLMs construct answers from parametric representations through probabilistic inference.
    A common implicit assumption is that a model knows a fact if it appeared in the pretraining data or if the model answers it correctly. However, prior work shows that both criteria are insufficient, as facts vary considerably in the strength of their encoding and the reliability of their recall~\cite{kandpal2023large, mallen2023not}.
    The inference pathway from input prompts through hidden representations and token distributions to generated answers ($x \to h \to p \to y$) therefore offers multiple points of intervention for improving knowledge extraction. Various inference-time techniques, which we group under \textit{Knowledge Extraction Enhancement (KEE)}, intervene at different stages of this pathway.
    Prompting methods modify the input context to elicit relevant knowledge~\cite{sun2023recitationaugmented, yu2023generate}, activation steering methods shift internal representations~\cite{li2023inference, rimsky-etal-2024-steering}, and decoding methods recalibrate output distributions~\cite{chuang2024dola, li2023contrastive}.
    Despite their methodological differences, these techniques share the goal of improving parametric knowledge extraction without updating the underlying model's parameters.

    Their evaluation, however, remains fragmented, as KEE methods typically reuse benchmarks designed for different objectives, with heterogeneous task formats and metrics~\cite{mallen2023not, lin2022truthfulqa, yang2018hotpotqa, thorne2018fever}.
    Following these benchmarks' original emphasis on task performance, KEE evaluations often focus on aggregate accuracy rather than on how interventions change individual predictions~\cite{kwiatkowski2019natural, joshi2017triviaqa, mallen2023not}.
    For instance, two methods can achieve the same accuracy while one largely preserves baseline answers and the other recovers many failures but disrupts many successes.
    These limitations call for a benchmark that supports consistent comparisons across KEE techniques and diagnoses their effects beyond aggregate accuracy. 
    
    We introduce \textbf{MAPLE}, a unified framework for diagnosing how KEE techniques change factual knowledge extraction relative to a model's own baseline (Figure~\ref{fig:intro}).
    MAPLE combines a standardized benchmark with multi-axis diagnostic metrics, providing a common basis for comparing KEE techniques.
    At the output level, we track correctness changes for each question, separately quantifying the recovery of baseline failures and the disruption of baseline successes.
    This separation reveals how much a technique recovers and what it sacrifices, exposing trade-offs that a single accuracy score can conceal.
    Beyond these correctness transitions, we also examine questions that remain incorrectly answered by tracking how the rank of the first gold token shifts between baseline and KEE technique. 
    This probability level reveals whether the gold token is ranked higher, equal or lower under the KEE technique, which is invisible at the surface level. Moreover, prior evaluations rarely ask whether a technique recovers answers beyond what the model already reaches through its own sampling. Since sampling requires no intervention, a KEE technique offers a real benefit when it recovers answers that sampling alone cannot reach. We further improve the reliability of our benchmark by verifying the presence of the queried facts in pretraining data, reducing the risk of conflating a lack of prior exposure with the failure of a KEE technique.
    To this end, we construct the dataset from a pre-cutoff Wikipedia dump and retain only facts whose sentences appear verbatim in the pretraining corpus of OLMo2~\cite{olmo20242}, for which training data is publicly available.
    Although corpus presence does not guarantee successful encoding or extractability, this verification establishes that OLMo2 had an opportunity to learn each fact, and it serves as a proxy for the other models that share the same Wikipedia source.

    We evaluate six KEE techniques across three model families (OLMo2, Qwen2.5, and Llama3) using \method.
    This unified comparison reveals distinct patterns that aggregate accuracy alone cannot explain.
    Among the evaluated techniques, prompting methods provide the strongest recovery, including answers missed by fixed-budget sampling, but also disrupt previously correct responses.
    ITI remains close to the baseline across the diagnostic axes, whereas the evaluated decoding methods and CAA alter gold-token ranks on unrecovered failures without consistently improving them.
    The contrast between ITI and CAA further shows that methods acting at the same stage of inference need not produce similar effects.
    Together, these findings demonstrate the value of comparing KEE techniques under a shared diagnostic framework, rather than treating aggregate accuracy as a complete account of their effects on knowledge extraction.

    Our contributions are threefold:
    \begin{itemize}
        \item \textbf{A standardized factual-recall benchmark} that provides a common evaluation task across KEE families, with verified pretraining-corpus presence for OLMo2.
        \item \textbf{A multi-axis diagnostic protocol} that separates answer recovery and disruption, gold-token rank shifts on unrecovered failures, and recovery relative to fixed-budget sampling.
        \item \textbf{A unified empirical analysis} across six techniques and three model families, revealing differences within and across KEE families that are obscured by aggregate accuracy.
    \end{itemize}
\section{Related Work}
\label{sec:sec2}

\paragraph{KEE Techniques.} 
Numerous methods aim to improve parametric knowledge extraction without additional training, but they span diverse tasks and methodological approaches. Decoding techniques recalibrate output distributions using contrastive signals from amateur counterparts, such as smaller models (Contrastive Decoding; \citealp{li2023contrastive}), early layers (DoLa; \citealp{chuang2024dola}), masked attention heads (DeCoRe; \citealp{gema-etal-2025-decore}), or activation entropy (Activation Decoding; \citealp{pmlr-v235-chen24av}). Steering techniques shift internal activations toward truthful directions, using steering vectors obtained from linear probes (ITI; \citealp{li2023inference}), contrastive examples (CAA; \citealp{rimsky-etal-2024-steering}), or input-adaptive semantics (SADI; \citealp{wang2025semantics}).
Prompting techniques take a two-stage approach, first eliciting the model's implicit knowledge as explicit text, then conditioning the final generation on this self-produced context, using either demonstrations (RECITE; \citealp{sun2023recitationaugmented}) or zero-shot instructions (GenRead; \citealp{yu2023generate}). 

\paragraph{Knowledge and Factuality Benchmarks.} Existing benchmarks used to evaluate KEE techniques were originally designed for other purposes: TriviaQA \cite{joshi2017triviaqa} is constructed from open-web and Wikipedia sources; Natural Questions \cite{kwiatkowski2019natural} is built from Google search queries; HotpotQA \cite{yang2018hotpotqa} measures multi-hop reasoning; TruthfulQA \cite{lin2022truthfulqa} evaluates susceptibility to human misconceptions; FACTOR \cite{muhlgay2024generating} and FEVER \cite{thorne2018fever} assess the capability to choose the factually correct option from explicitly given candidates, rather than generating the answer from the model’s own knowledge; and PopQA \cite{mallen2023not} focuses narrowly on long-tail entities rather than balanced knowledge distributions. 

\section{\method}
\label{sec:sec3}

\method\ comprises two components: an exposure-controlled benchmark (\S\ref{sec:sec3.1}) and a set of diagnostic metrics (\S\ref{sec:sec3.3}) that characterize how a KEE technique changes model predictions.

\subsection{Benchmark Construction}
\label{sec:sec3.1}
We construct our benchmark through a three-stage pipeline. 
We first extract knowledge triplets, the atomic units of factual knowledge, from a pre-cutoff Wikipedia dump. 
We then apply multi-stage filtering to retain triplets that are valid and unambiguous while preserving broad knowledge coverage. 
Finally, we verify the presence of each remaining triplet in OLMo2's open pretraining corpus to further strengthen its reliability by reducing the risk of conflating a lack of prior exposure with ineffective knowledge extraction.

\paragraph{Stage 1: Knowledge Triplet Extraction.}  
Following \citet{shi2024detecting, zhang2025min, fu2025mia}, we assume that Wikipedia data predating a model's training cutoff has been seen during the model's pretraining. As KEE techniques aim to recover knowledge acquired during pretraining, we leverage this assumption by constructing our benchmark from a March 2022 Wikipedia dump, ensuring temporal precedence over the training cutoffs of all evaluated models. Temporal precedence alone does not guarantee that a specific fact appeared in a model's pretraining data, so we directly verify corpus presence in Stage 3.  Since a single text can encompass multiple facts of heterogeneous extraction difficulty, document- or passage-level evaluation is too coarse. We therefore decompose the unstructured corpora into atomic (subject, relation, object) triplets using REBEL \cite{cabot2021rebel}, a seq2seq relation-extraction model. Finally, to ensure broad topical coverage, we map document entities to Wikidata's \textit{instance of} properties, organizing them into a custom taxonomy of ten domains (e.g., science, history, law). Table~\ref{tab:categories} shows the full domain categories with representative \textit{instance of} examples. Before extraction, we remove list-style articles, HTML artifacts, and documents under 100 words from the source corpus.

\paragraph{Stage 2: Multi-stage Filtering.}
Raw triplets from Stage 1 often contain noise and exhibit skewed distributions. To ensure validity, we map entities and relations to Wikidata identifiers using Wikimapper and verify their existence via the Wikidata API. We discard instances with domain mismatches via LLM-based filtering\footnote{We used gpt-4o-mini for all LLM-as-a-judge.}; specifically, we remove overly generic triplets that do not reflect their document's assigned domain. For quality control, we eliminate triplets containing uninformative relations (e.g., \textit{opposite of}, \textit{different from}) or direct lexical overlap between the subject and the object. Crucially, we restrict our benchmark strictly to single-answer triplets using the  Wikidata SPARQL API and LLM-based classification; multi-answer cases introduce evaluation ambiguity by conflating a model's lack of knowledge with its preference for an alternative valid entity.
In order to label entity frequency, we utilize monthly Wikipedia pageviews\footnote{Aggregated over March 2021--February 2022.} as a proxy for pretraining exposure. Using percentile thresholds, we assign each entity to a \textit{head}, \textit{torso}, or \textit{tail} bin, incorporating 20\% buffer zones between adjacent bins (10\% \text{head}, 40\% \text{torso}, 10\% \text{tail}, separated by 20\% buffer margins) to ensure that frequency bins are separated by distinct margins rather than arbitrary cutoffs. Finally, to prevent specific relation dominance, we cap each relation at 300 instances, yielding a candidate pool that spans a broad frequency distribution. This pool consists of 3,254 triplets in a short-answer question format, which we further verify against pretraining data in Stage 3. See Appendix~\ref{sec:appA} for benchmark construction details.

\paragraph{Stage 3: Corpus Presence Filtering.}


Temporal precedence in Stage~1 does not establish whether individual facts appeared in pretraining data.
We therefore filter the remaining triplets against OLMo-mix-1124~\cite{olmo20242}, OLMo2's publicly available pretraining corpus, through the following three steps.

\subparagraph{Candidate sentences.}
We map each triplet $(s, r, o)$ from Section~\ref{sec:sec3.1} to the Wikipedia document from which REBEL extracted it and split the document into sentences. A sentence becomes a candidate when it mentions both the subject and the object, where we also match their Wikidata aliases. We match the subject and object rather than the relation because REBEL outputs relations from a fixed set of about 200 labels, which rarely appear verbatim in the text.

\subparagraph{Relation verification.}
A sentence that mentions both entities does not necessarily express the target relation, since the same pair can be linked by different relations. We therefore feed each candidate sentence back into REBEL and keep it as a supporting sentence only if REBEL re-extracts the same triplet $(s, r, o)$ with aliases taken into account. This step ensures that each supporting sentence expresses the target triplet rather than merely mentioning both entities.

\subparagraph{Corpus presence.}
We query each supporting sentence against OLMo-mix-1124 with exact matching through the \texttt{infini-gram} API \cite{liu2024infini} and record its count. A triplet is retained if at least one of its supporting sentences appears in the corpus. Of the 3,254 candidate triplets, 1,303 pass all three steps and form our final benchmark. The filter is conservative, as a removed triplet may still appear in the corpus in a paraphrased form or in a sentence our pipeline does not recover.

\subsection{Evaluation Metrics}
\label{sec:sec3.3}
We introduce three complementary evaluation metrics that systematically capture changes at the surface, probability, and beyond-sampling levels. Surface level computes  \textit{Gain} and \textit{Loss} based on generated outputs of models. The probability level captures internal dynamics via \textit{Probability Rank Shift (PRS)}, which measures how a KEE technique moves the rank of the gold token on failures that remain unrecovered. The beyond-sampling level compares each technique with temperature sampling through the \textit{Shallow Gain Rate (SGR)} and \textit{Deep Gain Rate (DGR)}. Together, these three dimensions constitute a diagnostic evaluation protocol of \method.

\paragraph{Surface Level.}
At the surface level, we evaluate the model using its actual generated outputs. Traditional accuracy merges the gain and loss resulting from interventions of KEE techniques into a single value. However, since they directly alter the model behavior, a metric that decouples improvement and degradation is required to isolate each effect. Thus, we define \textit{Gain} and \textit{Loss} as follows: Gain is the proportion of baseline failures that the KEE technique answers correctly (Eq~\ref{eq:eq1}); Loss is the proportion of baseline successes that the KEE technique answers incorrectly (Eq~\ref{eq:eq2}).

\begin{equation}
\text{Gain} = \frac{|S_{\text{KEE}} \cap F_{\text{base}}|}{|F_{\text{base}}|}
\label{eq:eq1}
\end{equation}

\begin{equation}
\text{Loss} = \frac{|F_{\text{KEE}} \cap S_{\text{base}}|}{|S_{\text{base}}|}
\label{eq:eq2}
\end{equation}
Here, $S_{\text{KEE}}$ and $F_{\text{KEE}}$ denote the sets of instances answered correctly and incorrectly under the KEE technique, respectively, while $S_{\text{base}}$ and $F_{\text{base}}$ represent those under the baseline model. By setting the denominators of \text{Gain} and \text{Loss} to $\vert{}F_{\text{base}}\vert{}$ and $\vert{}S_{\text{base}}\vert{}$ rather than the total dataset size, we measure recovery and degradation of correct answers.

\paragraph{Probability Level.}
Surface level evaluation only captures instances whose correctness flips between the baseline and a KEE technique. We therefore trace the internal dynamics at the probability level for the unflipped $0\to0$ instances that remain incorrect under both settings. We refer to these instances as \textit{unrecovered failures}. Prior work evaluates factual knowledge by how highly a model ranks the gold token \cite{petroni2019language}. We extend this view from static presence to paired change and introduce \textit{Probability Rank Shift (PRS)}, which measures how a KEE technique moves the rank of the gold token on the same fact even when a failure does not turn into a success.

We first compute the probability of the full gold answer with teacher forcing and observe that once the first token is given, the remaining tokens become nearly deterministic and add little information beyond it. We therefore use only the first token, which is predicted under the same condition as free generation. Let $r_{\text{base}}$ and $r_{\text{KEE}}$ denote the rank of the first gold token over the full vocabulary under the baseline and the KEE technique. We exclude instances where the gold token is ranked first under either setting, since their failure lies beyond the first token. Then, we assign each remaining instance to one of three categories:
\begin{equation}
c =
\begin{cases}
\text{Improved} & \text{if } r_{\text{KEE}} < r_{\text{base}} \\
\text{Tied} & \text{if } r_{\text{KEE}} = r_{\text{base}} \\
\text{Worsened} & \text{if } r_{\text{KEE}} > r_{\text{base}}
\end{cases}
\label{eq:prs}
\end{equation}
Improved, Tied and Worsened indicate that the gold token is ranked higher, equal or lower under the KEE technique than under the baseline. We then report PRS as the proportion of each category:
\begin{equation}
\text{PRS}_{c} = \frac{N_{c}}{N_{\text{Improved}} + N_{\text{Tied}} + N_{\text{Worsened}}}
\label{eq:prs_ratio}
\end{equation}
where $N_{c}$ is the number of instances in category $c \in \{\text{Improved}, \text{Tied}, \text{Worsened}\}$.

\paragraph{Beyond-Sampling Level.}
Beyond-sampling level analysis asks whether a KEE technique recovers instances that the model fails to reach through its own sampling. As a point of comparison, we use temperature sampling at $\tau=0.7$ ($\tau_{0.7}$) and $\tau=1.2$ ($\tau_{1.2}$), since changing the temperature alone can also yield gains, which allows us to test whether KEE gains reflect an actual effect beyond sampling noise. We first partition baseline failures by whether sampling reaches the correct answer, using an initial seed (Seed A = 42) for each temperature. Shallow failures $\mathcal{D}_{\text{shallow}}$ are incorrect under greedy decoding but reached at least once within $K$ samples, whereas deep failures $\mathcal{D}_{\text{deep}}$ are never reached within $K$ samples. Shallow and deep thus refer to sampling recoverability for each model rather than to the objective difficulty of a fact. We set $K=20$ by sweeping $K$ from 1 to 30. At $K=20$, every model already identifies about 90\% of the shallow failures found with 30 samples at both temperatures. To check that this choice does not drive our conclusions, we compare the ranking of methods at each $K$ with the ranking at $K=20$ using the Kendall rank correlation, which equals 1 when two rankings agree on the order of every pair of methods. The correlation stays at 0.89 or higher for both SGR and DGR at every $K$ in Table~\ref{tab:k_tau}, so the ranking of methods remains stable regardless of $K$. We then evaluate each method $m \in \{\text{KEE}, \tau_{0.7}, \tau_{1.2}\}$ on these subsets. For $\tau_{0.7}$ and $\tau_{1.2}$, we draw $K$ new samples with an independent seed (Seed B = 0) so that the samples used for partitioning are not reused for evaluation. Using the sizes of $\mathcal{D}_{\text{shallow}}$ and $\mathcal{D}_{\text{deep}}$ as denominators, we define the \textit{Shallow Gain Rate (SGR)} and \textit{Deep Gain Rate (DGR)} as Eq.~\ref{eq:eq5} and Eq.~\ref{eq:eq6}:

\begin{equation}
\text{SGR} = \frac{|S_{m} \cap \mathcal{D}_{\text{shallow}}|}{|\mathcal{D}_{\text{shallow}}|}
\label{eq:eq5}
\end{equation}

\begin{equation}
\text{DGR} = \frac{|S_{m} \cap \mathcal{D}_{\text{deep}}|}{|\mathcal{D}_{\text{deep}}|}
\label{eq:eq6}
\end{equation}
where $S_{m}$ is the set of instances recovered by method $m$. SGR measures how much a method recovers among failures that sampling can also recover, and DGR measures how much it recovers among failures that sampling cannot recover within $K$ samples. A KEE technique that reaches a DGR above that of temperature sampling therefore recovers failures beyond what sampling alone provides.

\begin{table}[t]
\centering
\small
\begin{tabular}{lcccc}
\toprule
 & \multicolumn{2}{c}{$\tau_{0.7}$} & \multicolumn{2}{c}{$\tau_{1.2}$} \\
\cmidrule(lr){2-3} \cmidrule(lr){4-5}
$K$ & SGR & DGR & SGR & DGR \\
\midrule
10 & 0.915 & 0.941 & 0.988 & 0.898 \\
15 & 0.927 & 0.969 & 0.958 & 0.956 \\
20 & 1.000 & 1.000 & 1.000 & 1.000 \\
25 & 0.983 & 0.991 & 0.972 & 0.950 \\
30 & 0.956 & 0.987 & 0.931 & 0.950 \\
\bottomrule
\end{tabular}
\caption{Mean Kendall rank correlation with the ranking at $K=20$.}
\label{tab:k_tau}
\end{table}
\section{Experiments}

\subsection{Settings}
\paragraph{Models.}

To examine the effectiveness of KEE techniques across different model scales and families, we conduct experiments on OLMo2-7B/32B-Instruct, Qwen2.5-7B/32B-Instruct, and Llama3-8B/70B-Instruct. 

\paragraph{KEE Techniques.}
As discussed in Section~\ref{sec:sec2}, KEE techniques span a wide range of approaches, making it impractical to cover all existing methods. We therefore select two representative techniques from each of the three KEE families that are commonly used as baselines in prior work. From the decoding techniques, we include \textbf{DeCoRe} and \textbf{Activation Decoding (AD)}; from the steering techniques, \textbf{ITI} and \textbf{CAA}; from the prompting techniques, \textbf{RECITE} and \textbf{GenRead}. In addition, we compare KEE techniques against \textbf{temperature sampling}, sampled at $(\tau_{0.7})$ and $(\tau_{1.2})$. Detailed configurations for each technique are provided in Appendix~\ref{sec:appB}. 

\paragraph{Generation Setting.}
For the no-intervention baseline and all KEE techniques, we use greedy decoding to eliminate sampling noise and ensure reproducibility. Using greedy decoding, any change in the top-1 prediction strictly reflects the intervention impact of the KEE technique, allowing us to precisely measure Gain and Loss. For temperature sampling, the final answer is determined via majority voting across 10 samples. See Appendix~\ref{sec:appC} for detailed prompts. 

\paragraph{Evaluation \& Correctness Metrics.}
We evaluate the techniques using the metrics introduced in Section~\ref{sec:sec3.3} and accuracy (Acc). For correctness evaluation, we treat both the gold answer and the aliases as valid answers. After normalization---removing special characters and whitespace---a prediction is judged correct if it exactly matches a valid answer or if a valid answer appears as a substring of the prediction.

\begin{table*}[t]
\centering
\footnotesize
\renewcommand{\arraystretch}{1.15}
\setlength{\tabcolsep}{6pt}
\resizebox{\textwidth}{!}{%
\begin{tabular}{>{\centering\arraybackslash}m{1.8cm} | l r r r | l r r r | l r r r}
\toprule
\textbf{Category} & \textbf{Model} & \textbf{Acc$\uparrow$} & \textbf{Gain$\uparrow$} & \textbf{Loss$\downarrow$}
& \textbf{Model} & \textbf{Acc$\uparrow$} & \textbf{Gain$\uparrow$} & \textbf{Loss$\downarrow$}
& \textbf{Model} & \textbf{Acc$\uparrow$} & \textbf{Gain$\uparrow$} & \textbf{Loss$\downarrow$} \\
\midrule
\rowcolor{blue!8}
& \textbf{OLMo2-7B}       & 46.89 & --    & --    & \textbf{Qwen2.5-7B}    & 43.13 & --    & --    & \textbf{Llama3-8B}      & 50.58 & --    & --    \\
\multirow{2}{*}{\textbf{Sampling}}
& \quad + $\tau_{0.7}$ & 47.35 & 3.32  & \textbf{2.78}  & \quad + $\tau_{0.7}$ & 43.21 & 1.35  & \textbf{1.60}  & \quad + $\tau_{0.7}$ & 50.88 & 2.95  & \textbf{2.28}  \\
& \quad + $\tau_{1.2}$ & 45.43 & 3.61  & 7.20  & \quad + $\tau_{1.2}$ & 42.98 & 3.51  & 4.98  & \quad + $\tau_{1.2}$ & 50.27 & 6.21  & 6.68  \\
\rowcolor{gray!8}
& \quad + DeCoRe & 46.89 & 3.32  & 3.76  & \quad + DeCoRe & 41.29 & 4.32  & 9.96  & \quad + DeCoRe & 53.34 & 11.18 & 5.46  \\
\rowcolor{gray!8}
\multirow{-2}{*}{\textbf{Decoding}}
& \quad + AD     & 46.12 & 1.88  & 3.76  & \quad + AD     & 40.45 & 3.78  & 11.21 & \quad + AD     & 52.11 & 10.09 & 6.83  \\
\multirow{2}{*}{\textbf{Steering}}
& \quad + ITI    & 47.89 & 4.91  & \underline{3.44}  & \quad + ITI    & 42.06 & 1.21  & 4.09  & \quad + ITI    & 50.96 & 4.66  & \underline{3.79}  \\
& \quad + CAA    & 46.12 & 3.03  & 5.07  & \quad + CAA    & 42.67 & 1.48  & \underline{3.02} & \quad + CAA    & 51.50 & 11.02 & 8.95 \\
\rowcolor{gray!8}
& \quad + RECITE & \textbf{56.25} & \textbf{23.27} & 6.38  & \quad + RECITE & \underline{50.96} & \underline{20.65} & 9.07  & \quad + RECITE & \underline{56.49} & \underline{19.57} & 7.44  \\
\rowcolor{gray!8}
\multirow{-2}{*}{\textbf{Prompting}}
& \quad + GenRead& \underline{52.65} & \underline{19.94} & 10.31 & \quad + GenRead& \textbf{52.03} & \textbf{24.97} & 12.28 & \quad + GenRead& \textbf{58.63} & \textbf{28.88} & 12.29 \\
\addlinespace[3pt]\midrule\addlinespace[1pt]
\rowcolor{blue!8}
& \textbf{OLMo2-32B}      & 63.24 & --    & --    & \textbf{Qwen2.5-32B}   & 56.10 & --    & --    & \textbf{Llama3-70B}     & 67.84 & --    & --    \\
\multirow{2}{*}{\textbf{Sampling}}
& \quad + $\tau_{0.7}$ & 62.24 & 3.13  & 3.40  & \quad + $\tau_{0.7}$ & 56.10 & 1.40  & \underline{1.09}  & \quad + $\tau_{0.7}$ & 67.77 & 1.67  & \textbf{0.90}  \\
& \quad + $\tau_{1.2}$ & 63.01 & 6.89  & 4.37  & \quad + $\tau_{1.2}$ & 56.49 & 3.32  & 1.92  & \quad + $\tau_{1.2}$ & 67.92 & 3.10  & 1.36  \\
\rowcolor{gray!8}
& \quad + DeCoRe & 63.55 & 4.80  & \underline{2.31}  & \quad + DeCoRe & 55.72 & 1.40  & 1.78  & \quad + DeCoRe & 68.53 & 5.49  & 1.58  \\
\rowcolor{gray!8}
\multirow{-2}{*}{\textbf{Decoding}}
& \quad + AD     & 62.32 & 3.13  & 3.28  & \quad + AD     & 56.25 & 1.05  & \textbf{0.55}  & \quad + AD     & 68.30 & 8.11  & 3.17  \\
\multirow{2}{*}{\textbf{Steering}}
& \quad + ITI    & 63.39 & 2.92  & \textbf{1.46}  & \quad + ITI    & 55.56 & 1.57  & 2.19  & \quad + ITI    & 67.92 & 2.86  & \underline{1.24}  \\
& \quad + CAA    & 62.70 & 5.85  & 4.25  & \quad + CAA    & 56.41 & 2.10  & \underline{1.09} & \quad + CAA    & 67.77 & 8.59  & 4.19  \\
\rowcolor{gray!8}
& \quad + RECITE & \textbf{69.38} & \textbf{26.93} & 5.95  & \quad + RECITE & \textbf{61.55} & \underline{22.03} & 7.52  & \quad + RECITE & \underline{70.76} & \underline{22.43} & 6.33  \\
\rowcolor{gray!8}
\multirow{-2}{*}{\textbf{Prompting}}
& \quad + GenRead& \underline{66.69} & \underline{23.38} & 8.13  & \quad + GenRead& \underline{59.71} & \textbf{22.90} & 11.49 & \quad + GenRead& \textbf{73.83} & \textbf{32.70} & 6.67  \\
\bottomrule
\end{tabular}
}
\caption{Gain/Loss (\%) decomposition across model families and KEE techniques. \textbf{Bold} and \underline{underline} indicate the best and second-best values, respectively. Alternating gray shading indicates technique category boundaries.}
\label{tab:tab1}
\end{table*}

\subsection{Main Results}
\label{sec:sec4.2}

Table~\ref{tab:tab1} reports accuracy together with Gain and Loss for each KEE technique where the temperature sampling serves as the reference. Figure~\ref{fig:prs} shows PRS on unrecovered failures for each KEE technique. Each bar and its value give the averaged PRS over the six models. Range bars mark the minimum and maximum of that PRS across the six models. Table~\ref{tab:tab2} reports SGR and DGR at $\tau=1.2$, which separate the gains of KEE technique on shallow failures that sampling reaches within $K$ samples from those on deep failures that sampling never reaches. Full results and raw count breakdown are in Appendix~\ref{sec:appD}.

\paragraph{A Diagnostic Lens with Three Complementary Metrics.}
Each metric in Section~\ref{sec:sec3.3} covers a blind spot that the others leave open. Gain and Loss decompose accuracy into the failures a technique recovers and the successes it breaks, recovering bidirectional shifts that a single accuracy score conflates. Since they register a change only when the answer flips, PRS complements them by tracing how the gold token's rank moves on unrecovered failures. Raising the sampling temperature alone can also improve correctness without any intervention, so SGR and DGR compare each technique with temperature sampling to test whether its recovery merely reflects what sampling can recover or goes beyond it. Together the metrics capture correctness flips in both directions, the dynamics beneath unflipped failures, and the margin over a temperature sampling. Because all of them rely only on answer flips or ranks, they apply to every KEE technique and remain comparable across decoding, steering and prompting techniques that reshape the output distribution in different ways.

\paragraph{What the Unified Lens Reveals about KEE Techniques.}
While the three metrics capture distinct dimensions, they are fundamentally interconnected. Probability level dynamics are reflected at the surface level, and a robust signal at the surface level accompanies the recovery of deep failures beyond sampling. Prompting techniques such as RECITE and GenRead exemplify this end-to-end connection. By injecting relevant context directly into the input, they reshape the model's probability distribution most decisively and leave only a small share of $\text{PRS}_{\text{Tied}}$, which coincides with both the largest Gain and the strongest recovery of deep failures. Steering via ITI lies at the opposite end. It shows the largest share of $\text{PRS}_{\text{Tied}}$ across models, and this minimal reshaping is mirrored by small Gain alongside low Loss and DGR close to zero, keeping ITI close to the baseline at every level. Decoding-based techniques such as DeCoRe and AD and steering via CAA fall in between. They reshape the distribution more than ITI, yet this reshaping is not directed toward the gold token as $\text{PRS}_{\text{Worsened}}$ exceeds $\text{PRS}_{\text{Improved}}$ on average. Correspondingly, their Gain remains comparable to that of temperature sampling in most models, and their knowledge recovery remains largely confined to shallow failures and rarely extends into deep ones.

\section{Analysis}
\label{sec:sec5}

\subsection{Surface Level}

\paragraph{Recalling knowledge makes substantial interventions.}
In Table~\ref{tab:tab1}, RECITE and GenRead achieve the highest Gain across all models, while GenRead also incurs the highest Loss in every model. RECITE keeps a comparably high Gain whereas its Loss stays lower than GenRead's in every model.

\paragraph{Decoding-based modifications produce comparatively weak interventions.}
In Table~\ref{tab:tab1}, decoding techniques provide weak interventions whose efficacy varies across model families. On Gain, they achieve 1.6--4.9$\times$ higher values than temperature sampling on the Llama3 family, but fall to 0.3--1.5$\times$ on the OLMo2 family and Qwen2.5-32B. On Loss, they remain comparable to temperature sampling and markedly lower than prompting techniques, except for a spike on Qwen2.5-7B where DeCoRe and AD reach 9.96\% and 11.21\%.

\paragraph{Activation steering behaviors diverge by specific methodology.}
CAA aligns more closely with decoding than with its fellow steering technique ITI on both Gain and Loss, exhibiting a similar model-dependent pattern. In Table~\ref{tab:tab1}, while CAA's Gain and Loss stay comparable to temperature sampling on OLMo2 and Qwen2.5, they reach 1.8--5.1$\times$ and 1.3--4.7$\times$ higher levels on the Llama3 family, respectively. Notably, on Qwen2.5 family, CAA shows both the lowest Gain at 1.48\% and 2.10\% and the lowest Loss at 3.02\% and 1.09\% among all six models, suggesting its intervention signal is comparatively muted for this model family. ITI produces a weak intervention signal and stays close to the baseline in every model, with Gain between 1.21\% and 4.91\% and consistently low Loss among KEE techniques across six models. Since prior decoding and steering literature has been validated predominantly on Llama \cite{gema-etal-2025-decore, rimsky-etal-2024-steering}, our findings put a spotlight on the evaluation of cross-family generalization as a critical direction for future KEE research.

\begin{figure}[t]
  \centering
  \includegraphics[width=0.49\textwidth]{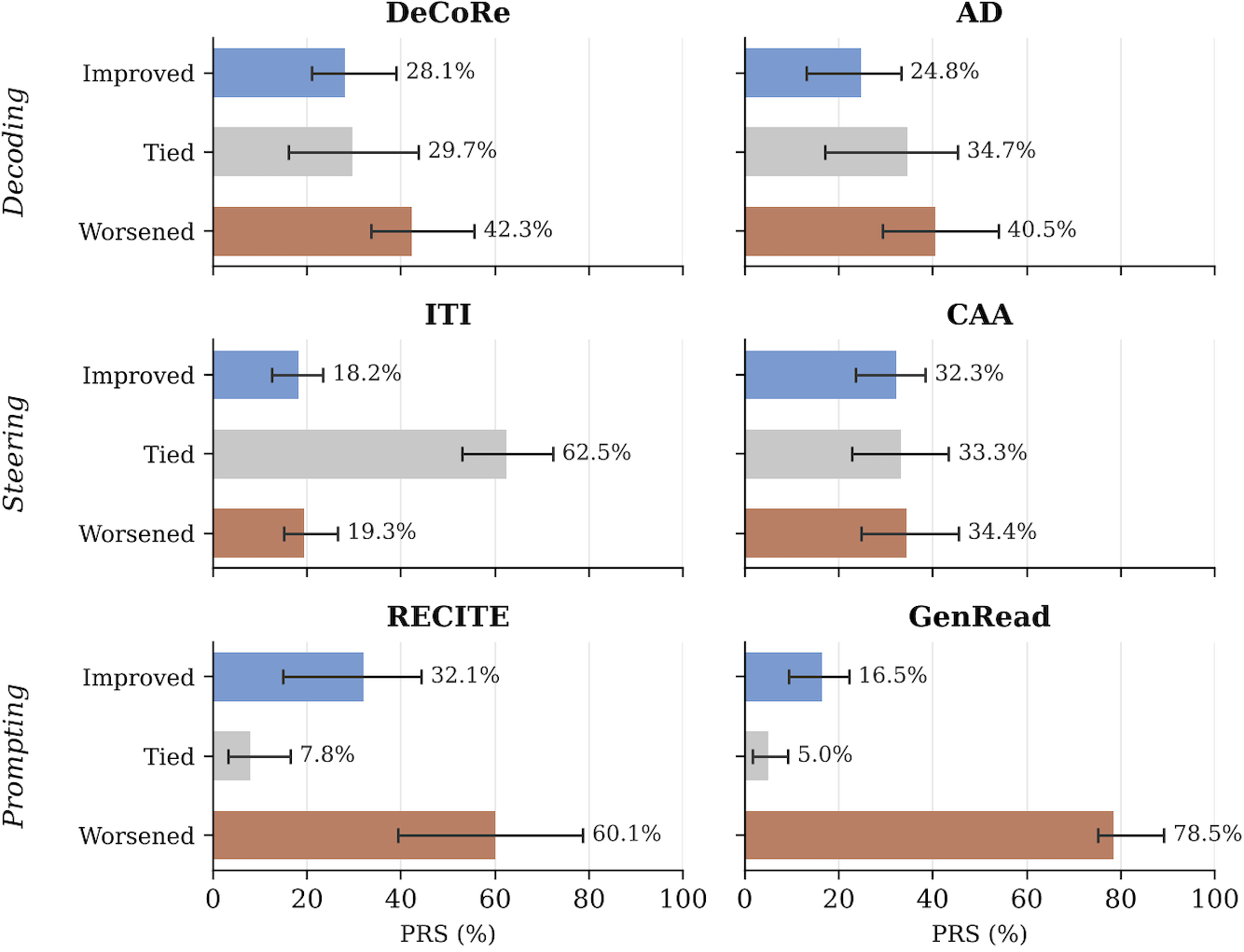}
    \caption{Probability Rank Shift of the gold token on unrecovered failures for KEE techniques, averaged over six models. Range bars show the minimum and maximum across models. Per-model results are in Figure~\ref{fig:prs_full}.}
  \label{fig:prs}
\end{figure}

\begin{table*}[t]
\centering
\footnotesize
\renewcommand{\arraystretch}{1.0}
\setlength{\tabcolsep}{8pt}
\resizebox{\textwidth}{!}{%
\begin{tabular}{>{\centering\arraybackslash}m{1.8cm} | l r r | l r r | l r r}
\toprule
\textbf{Category} & \textbf{Model} & \textbf{SGR$\uparrow$} & \textbf{DGR$\uparrow$}
& \textbf{Model} & \textbf{SGR$\uparrow$} & \textbf{DGR$\uparrow$}
& \textbf{Model} & \textbf{SGR$\uparrow$} & \textbf{DGR$\uparrow$} \\
\midrule
\rowcolor{blue!8}
& \textbf{OLMo2-7B}       & --    & --    & \textbf{Qwen2.5-7B}    & --    & --    & \textbf{Llama3-8B}      & --    & --    \\
\textbf{Sampling}
& \quad + $\tau_{1.2}$ & 16.06 & 0.42  & \quad + $\tau_{1.2}$ & 18.67 & 0.17  & \quad + $\tau_{1.2}$ & 15.72 & 0.24  \\
\rowcolor{gray!8}
& \quad + DeCoRe & 9.63  & 0.42  & \quad + DeCoRe & 16.00 & 1.35  & \quad + DeCoRe & 24.02 & 4.10  \\
\rowcolor{gray!8}
\multirow{-2}{*}{\textbf{Decoding}}
& \quad + AD     & 5.96  & 0.00  & \quad + AD     & 15.33 & 0.85  & \quad + AD     & 20.96 & 4.10  \\
\multirow{2}{*}{\textbf{Steering}}
& \quad + ITI    & 14.68 & 0.42  & \quad + ITI    & 6.00  & 0.00  & \quad + ITI    & 10.04 & 1.69  \\
& \quad + CAA    & 9.63  & 0.00  & \quad + CAA    & 6.67  & 0.17  & \quad + CAA    & 23.58 & 4.10  \\
\rowcolor{gray!8}
& \quad + RECITE & \textbf{51.83} & \textbf{10.13} & \quad + RECITE & \underline{53.33} & \underline{12.35} & \quad + RECITE & \underline{41.48} & \underline{7.47}  \\
\rowcolor{gray!8}
\multirow{-2}{*}{\textbf{Prompting}}
& \quad + GenRead& \underline{44.50} & \underline{8.65} & \quad + GenRead& \textbf{60.00} & \textbf{16.07} & \quad + GenRead& \textbf{52.84} & \textbf{15.66} \\
\addlinespace[3pt]\midrule\addlinespace[1pt]
\rowcolor{blue!8}
& \textbf{OLMo2-32B}      & --    & --    & \textbf{Qwen2.5-32B}   & --    & --    & \textbf{Llama3-70B}     & --    & --    \\
\textbf{Sampling}
& \quad + $\tau_{1.2}$ & 13.07 & 0.00  & \quad + $\tau_{1.2}$ & 14.55 & 0.22  & \quad + $\tau_{1.2}$ & 16.33 & 0.00  \\
\rowcolor{gray!8}
& \quad + DeCoRe & 10.55 & 0.71  & \quad + DeCoRe & 7.27  & 0.00  & \quad + DeCoRe & 18.37 & 1.56  \\
\rowcolor{gray!8}
\multirow{-2}{*}{\textbf{Decoding}}
& \quad + AD     & 7.04  & 0.36  & \quad + AD     & 5.45  & 0.00  & \quad + AD     & 26.53 & 2.49  \\
\multirow{2}{*}{\textbf{Steering}}
& \quad + ITI    & 7.04  & 0.00  & \quad + ITI    & 8.18  & 0.00  & \quad + ITI    & 11.22 & 0.31  \\
& \quad + CAA    & 14.07 & 0.00  & \quad + CAA    & 10.91 & 0.00  & \quad + CAA    & 29.59 & 2.18  \\
\rowcolor{gray!8}
& \quad + RECITE & \textbf{48.74} & \textbf{11.43} & \quad + RECITE & \textbf{51.82} & \underline{14.94} & \quad + RECITE & \underline{46.94} & \underline{14.95} \\
\rowcolor{gray!8}
\multirow{-2}{*}{\textbf{Prompting}}
& \quad + GenRead& \underline{43.22} & \underline{9.29} & \quad + GenRead& \underline{41.82} & \textbf{18.40} & \quad + GenRead& \textbf{59.18} & \textbf{24.61} \\
\bottomrule
\end{tabular}
}
\caption{SGR/DGR (\%) across model families and KEE techniques at $\tau{=}1.2$. \textbf{Bold} and \underline{underline} indicate the best and second-best values, respectively. Alternating gray shading indicates technique category boundaries.}
\label{tab:tab2}
\end{table*}

\subsection{Probability Level}

\paragraph{Unrecovered failures are reshaped without moving toward the gold answer.} For decoding techniques and CAA, although they seem ineffective at the surface level with low Gain, most of their unrecovered failures still change the gold rank. Only 29.7\%, 34.7\% and 33.3\% of these failures remain Tied for DeCoRe, AD and CAA in Figure~\ref{fig:prs}. This reshaping, however, is not directed toward the gold token. On average, $\text{PRS}_{\text{Worsened}}$ exceeds $\text{PRS}_{\text{Improved}}$ for all three techniques, and the gap is pronounced in some models. For instance, on Llama3-70B, AD shows a similar gap at 46.7\% against 13.1\% in Table~\ref{tab:prs_raw}. These results indicate that most unchanged cases are still affected, but the change rarely brings the gold answer closer to the top. 

\paragraph{ITI leaves most unrecovered failures untouched.} ITI shows the largest $\text{PRS}_{\text{Tied}}$ at 62.5\% in Figure~\ref{fig:prs}, ranging from 53.0\% to 72.2\% across models in Table~\ref{tab:prs_raw}. Its $\text{PRS}_{\text{Improved}}$ and $\text{PRS}_{\text{Worsened}}$ are nearly balanced at 18.2\% and 19.3\%, so the few changes it makes carry no consistent direction. This minimal reshaping is consistent with its small Gain and low Loss at the surface level in Table~\ref{tab:tab1}.

\paragraph{Context injection strongly reshapes the output distributions.} Both prompting techniques show the smallest $\text{PRS}_{\text{Tied}}$ at about 7.8\% for RECITE and 5.0\% for GenRead in Figure~\ref{fig:prs}. However, their consistency differs. GenRead consistently pushes the gold token downward, with $\text{PRS}_{\text{Worsened}}$ ranging from 75.1\% to 89.1\% across models. In contrast, RECITE exhibits less consistency across model families, with its $\text{PRS}_{\text{Improved}}$ ranging from 14.8\% on OLMo2-32B to 44.3\% on Llama3-70B in Table~\ref{tab:prs_raw}.

\subsection{Beyond-Sampling Level}
\label{sec:sec5.3}

\paragraph{Weak interventions mostly recover shallow failures.} In Tables~\ref{tab:tab2} and~\ref{tab:sampling_t07}, decoding techniques and CAA demonstrate gains that are largely concentrated in shallow failures rather than deep ones. On OLMo2 and Qwen2.5 families, these interventions generally perform comparably to or lag behind $\tau_{0.7}$ and $\tau_{1.2}$. For instance, OLMo2-7B's AD achieves an SGR of 5.96\% and a DGR of 0.00\%, whereas $\tau_{1.2}$ reaches 16.06\% and 0.42\% in Table~\ref{tab:tab2}.  In contrast, the Llama3 family retains a clear advantage at both temperatures. Llama3-8B's CAA achieves a DGR of 4.10\%, far outperforming $\tau_{1.2}$ at 0.24\%. These interventions primarily recover shallow failures within the reach of temperature sampling rather than addressing deep failures that lie beyond sampling, with an exception on Llama3 where a substantial gap remains. ITI also stays at or below 2.06\% DGR in every model at both temperatures.

\paragraph{Prompting unlocks deep failures robustly across temperatures.} RECITE and GenRead decisively outperform $\tau_{0.7}$ and $\tau_{1.2}$ on both SGR and DGR. For instance, on OLMo2-32B, prompting techniques achieve a DGR of up to 15.76\% at $\tau=0.7$ and 11.43\% at $\tau=1.2$, compared to 0.00\% from both $\tau_{0.7}$ and $\tau_{1.2}$ in Tables~\ref{tab:tab2} and~\ref{tab:sampling_t07}.

\section{Conclusion}
We introduce \method, a unified framework for evaluating KEE techniques that comprises a benchmark of facts verified to be present in OLMo2's pretraining corpus and a diagnostic protocol spanning surface, probability, and beyond-sampling level metrics. Across three model families and six KEE techniques, \method\ reveals behaviors that aggregated accuracy cannot surface. Decoding techniques and CAA reshape the output distribution without moving it toward the gold answer and largely remain within the reach of temperature sampling, ITI stays close to the baseline at every level, and prompting techniques recover the most failures beyond sampling but also incur the largest Loss. By capturing these distinct behaviors across complementary axes, our framework offers a fine-grained lens to dissect the trade-offs of KEE techniques and a multi-axis standard to guide the development of more effective and stable methods for future LLMs.

\clearpage

\section*{Acknowledgements}
This research was supported by grants from the National Research Foundation of Korea (NRF), funded by the Korean government (MSIT and MOE: RS-2025-16063688; MSIT: RS-2025-02215813, RS-2026-25491306, RS-2025-25442252).


\section*{Limitations}
While our framework provides a systematic diagnostic lens for KEE techniques, several limitations remain.

\paragraph{Corpus Verification.}
We verify that every benchmark fact has a supporting sentence that appears verbatim in OLMo2's pretraining corpus and from which the same triplet is re-extracted as described in Stage 3 of Section~\ref{sec:sec3.1}. This verification establishes presence in pretraining data rather than encoding in model parameters. Encoding strength varies continuously with exposure \cite{kandpal2023large, mallen2023not}, so a fact present in the corpus may still be encoded only weakly or not at all. The verification is also limited to OLMo2, whose pretraining corpus is public. For Qwen2.5 and Llama3, we rely on the facts being drawn from a Wikipedia dump that predates their training cutoffs, which makes their presence likely but not verified. 

\paragraph{Model \& KEE Technique Coverage.}
Our benchmark, built from a March 2022 Wikipedia dump, applies only to models whose pretraining post-dates this cutoff. Due to computational constraints, we evaluate six models from three families, restricting the transferability of our findings to other families. Similarly, we evaluate two representative techniques from each of the three KEE technique categories (decoding, steering, prompting) rather than the full space of methods. Our framework, however, is model- and technique-agnostic and can be readily extended along both axes. We leave further investigation of other models and KEE techniques as future work.

\paragraph{Scope of Knowledge.}
Although our benchmark covers a wide range of domains, frequency levels, and relations, it only consists of factual triplets with a single answer, drawn from Wikipedia. There are different types of knowledge relevant to our work, such as ambiguous multi-answer facts or non-English knowledge. We do not cover these in this paper as they are out of our scope. Extending the benchmark to these settings will be an interesting direction for future work.

\bibliography{reference} 

\appendix
\clearpage

\section{Benchmark Construction Details}
\label{sec:appA}

We organize source entities into ten categories; the category set and
representative \textit{instance of} (\texttt{P31}) types are listed in Table~\ref{tab:categories}.

\subsection{Domain Classification Prompt}
\label{sec:domain_prompts}

We classify each triplet into one of ten domains using an LLM-based relevance
classifier. All ten domains share an identical system prompt template; only the domain
name (i.e., \texttt{<DOMAIN>}) and its entity set of in-domain scope (i.e., \texttt{<SCOPE>}) differ. The shared template is shown
in Figure~\ref{fig:domain_classification_prompt}, where \texttt{<DOMAIN>} and
\texttt{<SCOPE>} are instantiated per domain using the values listed in
Table~\ref{tab:domain_scopes}.

\subsection{Answer Cardinality Filtering Prompt}
\label{sec:cardinality_prompt}

To restrict the benchmark to triplets with a single valid answer, we filter each
candidate triplet with an LLM-based cardinality classifier. The classifier
labels a triplet as \texttt{SINGLE}, \texttt{MULTIPLE}, or \texttt{UNSURE}, and
only \texttt{SINGLE} triplets are retained. The system prompt is shown in
Figure~\ref{fig:cardinality_system}, and the few-shot demonstrations supplied to
the classifier are listed in Table~\ref{tab:cardinality_fewshot}.

\subsection{Benchmark Statistics}
\label{sec:benchmark_dist}

\begin{figure*}[t]
  \centering
  \includegraphics[width=\textwidth]{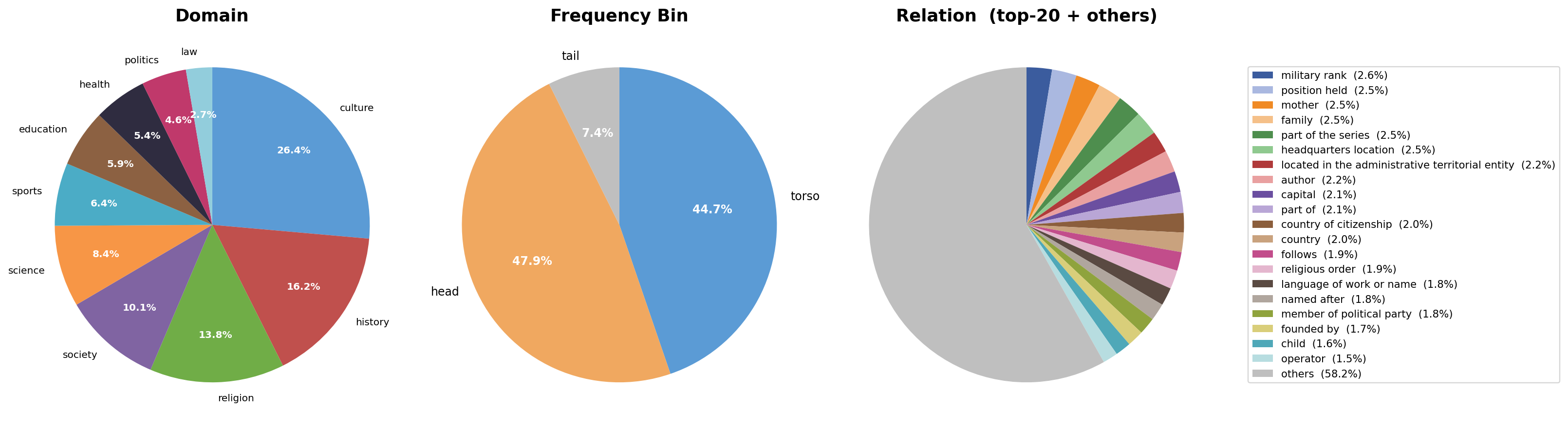}
  \caption{Distribution of the 1{,}303 triplets across domain, frequency bin,
  and relation, showing broad knowledge coverage along all three axes. Bins are
  defined by frequency percentiles rather than the targeted proportions; the tail bin
  is smaller (7.4\%) because low-frequency entities yield fewer triplets under
  our relation and domain constraints. The relation chart shows the top-20
  of 97 unique types.}
  \label{fig:dist}
\end{figure*}

Statistics of our benchmark are shown in Figure~\ref{fig:dist}.



\begin{table*}[t]
\centering
\renewcommand{\arraystretch}{1.5}
\begin{tabular}{>{\centering\arraybackslash}m{3.5cm}>{\raggedright\arraybackslash}m{11cm}}
\toprule
\textbf{Category} & \textbf{Instance of (\texttt{P31})} \\
\midrule
Culture & literary work, museum, art movement, heritage site, cultural center \\
\midrule
Education & school, university, college, campus, faculty \\
\midrule
Health & hospital, anatomical structure, medical treatment, disease, symptom or sign \\
\midrule
History & battle, war, dynasty, historical period, invasion \\
\midrule
Law & legal case, prison, court, criminal case, decision of the Supreme Court of the United States \\
\midrule
Politics & political organization, electoral unit, public election, constituency of the European Parliament, executive branch \\
\midrule
Religion & mythological Greek character, religious community, Catholic religious institute, Jewish organization, Islamic term \\
\midrule
Science & protein, mutant, spiral galaxy, chemical element, scientific theory \\
\midrule
Society & voluntary association, ethnic group, clan, human population, women's organization \\
\midrule
Sports & roller derby league, amateur football club, esports team, championship, rugby league venue \\
\bottomrule
\end{tabular}
\caption{Categories with representative \textit{instance of} (\texttt{P31}) types.}
\label{tab:categories}
\end{table*}
\begin{figure}[t]
\centering
\begin{tcolorbox}[
  colback=gray!5,
  colframe=gray!50,
  fonttitle=\bfseries,
  title=Domain Classification Prompt Template,
  boxrule=0.5pt,
  left=4pt, right=4pt, top=4pt, bottom=4pt,
]
\small\ttfamily
You are a domain relevance classifier for knowledge graph triplets.\\
Given a triplet (subject, relation, object), determine if it belongs to the <DOMAIN> domain.\\[4pt]
<DOMAIN> includes: <SCOPE>.\\[4pt]
Rule:\\
Classify as "<DOMAIN>" if any of the following is true:\\
- The subject or object is a <DOMAIN>-related entity.\\
- The triplet describes a <DOMAIN>-related fact.\\
Classify as "non-<DOMAIN>" only if none of the above conditions are met.\\[4pt]
Respond with ONLY "<DOMAIN>" or "non-<DOMAIN>".
\end{tcolorbox}
\caption{System prompt for the domain classifier.}
\label{fig:domain_classification_prompt}
\end{figure}
\begin{table*}[t]
\centering
\small
\begin{tabularx}{\textwidth}{@{}l X@{}}
\toprule
\textbf{Domain} & \textbf{In-domain scope (\texttt{<SCOPE>})} \\
\midrule
Culture &
literature, books, authorship, visual arts, paintings, sculpture, architecture,
performing arts, dance, museums, libraries, collections of cultural works,
monuments, cultural heritage, cultural institutions, festivals, and directly
related cultural concepts \\
\addlinespace
Education &
schools, universities, colleges, faculties, academic departments, school
districts, educational institutions, academic programs, degrees, curricula,
teachers, students, education systems, and directly related educational
concepts \\
\addlinespace
Health &
diseases, symptoms, anatomy, organs, mental health, bacteria, viruses, drugs,
vaccines, medicines, treatments, antibiotics, medical procedures, diagnosis,
healthcare institutions, hospitals, healthcare organizations, doctors, nurses,
medical specialties, and any entity or fact closely associated with these
topics \\
\addlinespace
History &
historical events, wars, battles, conflicts, military, empires, dynasties,
reigns, historical periods, historical rulers, archaeological periods, and
directly related historical concepts \\
\addlinespace
Law &
legal cases, court decisions, laws, legal concepts, crimes, courts, judicial
institutions, judges, legal offices, constitutions, and directly related legal
concepts \\
\addlinespace
Politics &
political parties, elections, governments, parliaments, ministries, presidents,
prime ministers, politicians, legislation, political institutions, protests,
revolution, political ideologies, diplomacy, international relations, and
directly related political concepts \\
\addlinespace
Religion &
religions, religious organizations, churches, temples, mosques, monasteries,
religious buildings, deities, mythological figures, mythology, rituals,
doctrines, saints, sacred places, and directly related religious concepts \\
\addlinespace
Science &
biology, species, taxonomy, genes, astronomy, planets, stars, galaxies,
chemistry, chemical elements, compounds, reactions, earth science, scientific
theories, scientific laws, scientific methods, scientists, research
institutions, and directly related scientific concepts \\
\addlinespace
Society &
social groups, ethnic groups, demographic groups, communities, tribes,
non-profit organizations, social movements, social issues, and directly related
societal concepts \\
\addlinespace
Sports &
sports teams, athletes in sports roles, leagues, tournaments, seasons, matches,
races, competitions, sporting events, stadiums, arenas, sports venues, coaches,
and directly related sports concepts \\
\bottomrule
\end{tabularx}
\caption{Per-domain instantiation of the prompt template in
Figure~\ref{fig:domain_classification_prompt}. Each row provides the
\texttt{<DOMAIN>} name and the \texttt{<SCOPE>} string substituted into the
shared template.}
\label{tab:domain_scopes}
\end{table*}
\begin{figure}[t]
\centering
\begin{tcolorbox}[
  colback=gray!5,
  colframe=gray!50,
  fonttitle=\bfseries,
  title=Answer Cardinality Classification Prompt Template,
  boxrule=0.5pt,
  left=4pt, right=4pt, top=4pt, bottom=4pt,
]
\small\ttfamily
You are given a factoid question.\\[4pt]
Determine whether this question has exactly one valid answer, or multiple valid answers.\\[4pt]
- SINGLE: The question has exactly one correct answer.\\
- MULTIPLE: The question has more than one valid answer. This includes:\\
\hspace*{2em}(a) Factually multiple answers: the entity genuinely has multiple valid values (e.g., a drug treats multiple diseases, a person held multiple positions, a country borders multiple neighbors).\\
\hspace*{2em}(b) Temporally variable answers: the answer changes over time, so multiple answers have been correct at different points (e.g., headquarters that relocated, capitals that changed, leaders who were replaced).\\
\hspace*{2em}(c) Perspectival ambiguity: the answer depends on which context (e.g., classification system, hierarchical system) is used.\\
- UNSURE: You are not confident enough to judge, or you do not know enough about the subject.\\[4pt]
Rule:\\
- Even if one answer is most famous or commonly cited, if other valid answers exist, classify as MULTIPLE.\\[4pt]
Output JSON:\\
\{"cardinality2": "SINGLE" or "MULTIPLE" or "UNSURE", "reason": str\}
\end{tcolorbox}
\caption{System prompt for the answer cardinality classifier.}
\label{fig:cardinality_system}
\end{figure}
\begin{table*}[t]
\centering
\small
\begin{tabularx}{\textwidth}{@{}l l l c X@{}}
\toprule
\textbf{Subject} & \textbf{Relation} & \textbf{Object} & \textbf{Label} & \textbf{Reason} \\
\midrule
Italy & capital & Rome & \texttt{SINGLE} &
Italy has exactly one capital, Rome, which has not changed. \\
\addlinespace
aspirin & medical condition treated & pain & \texttt{MULTIPLE} &
Aspirin treats multiple conditions including pain, fever, inflammation, and
cardiovascular disease. \\
\addlinespace
Microsoft & headquarters location & Redmond & \texttt{MULTIPLE} &
The headquarters of Microsoft has changed over time (Albuquerque, then
Redmond), so the answer depends on the time period. \\
\addlinespace
Winston Churchill & conflict & World War II & \texttt{MULTIPLE} &
Churchill participated in multiple conflicts including the Second Boer War,
World War I, and World War II. \\
\addlinespace
amygdala & part of & limbic system & \texttt{SINGLE} &
The amygdala is part of exactly one brain system: the limbic system. \\
\addlinespace
Kaba-class destroyer & country & Japan & \texttt{UNSURE} &
Not confident enough about this specific naval vessel class to determine
whether it was operated by one or multiple countries. \\
\bottomrule
\end{tabularx}
\caption{Few-shot demonstrations provided to the answer cardinality classifier.
Each demonstration pairs an input triplet with its gold label and rationale.}
\label{tab:cardinality_fewshot}
\end{table*}

\section{Experiments Details}
\label{sec:appB}

We present the configurations we applied to implement KEE techniques. 
Unless otherwise noted, hyperparameters follow the recommendations of each
method's original paper.

\subsection{Decoding Methods}
\label{sec:hp_decoding}

\paragraph{DeCoRe.}
We set \texttt{num\_retrieval\_head}\,$=10$ for all models, following the original paper, which reports that performance on factuality benchmarks degrades as more retrieval heads are masked \cite{gema-etal-2025-decore}. As a post-hoc check in our setting, Table~\ref{tab:decore_sweep} shows that performance is nearly identical between 10 and 20 heads and degrades as more heads are masked.

\begin{table}[t]
\centering
\resizebox{\columnwidth}{!}{%
\begin{tabular}{lccccc}
\toprule
\multirow{2}{*}{\textbf{Model}} & \multicolumn{5}{c}{\texttt{num\_retrieval\_head}} \\
\cmidrule(l){2-6}
 & \textbf{10} & \textbf{20} & \textbf{30} & \textbf{40} & \textbf{50} \\
\midrule
Llama3-8B-Instruct  & 0.5236 & 0.5265 & 0.5178 & 0.5170 & 0.5167 \\
Qwen2.5-7B-Instruct & 0.4017 & 0.4028 & 0.3981 & 0.3899 & 0.3748 \\
\bottomrule
\end{tabular}%
}
\caption{DeCoRe accuracy on our benchmark as \texttt{num\_retrieval\_head} varies from 10 to 50. Unlike the main results, which use greedy decoding, accuracy here is the mean per-fact correctness over $N=10$ sampled trials at temperature $T=0.7$.}
\label{tab:decore_sweep}
\end{table}

\paragraph{Activation Decoding.}
The original paper~\cite{pmlr-v235-chen24av} adopts the layer 26 for a 32-layer model---i.e., $81.25\%$ of the total depth---for its main
experiments. We therefore set \texttt{info\_layer} to the layer at $81.25\%$ of
each model's total depth, as listed in Table~\ref{tab:hp_activation_decoding}. Also, we set $\alpha=0.5$, which controls the degree of entropy.

\begin{table}[t]
\centering
\small
\begin{tabular}{@{}lc@{}}
\toprule
\textbf{Model} & \texttt{info\_layer} \\
\midrule
OLMo2-7B-Instruct   & 26 / 32 \\
OLMo2-32B-Instruct  & 52 / 64 \\
Qwen2.5-7B-Instruct   & 23 / 28 \\
Qwen2.5-32B-Instruct  & 52 / 64 \\
Llama3-8B-Instruct   & 26 / 32 \\
Llama3-70B-Instruct   & 65 / 80 \\
\bottomrule
\end{tabular}
\caption{Per-model \texttt{info\_layer} for Activation Decoding.}
\label{tab:hp_activation_decoding}
\end{table}
\begin{table}[t]
\centering
\small
\begin{tabular}{@{}lcc@{}}
\toprule
\multirow{2}{*}{\textbf{Model}} & \multicolumn{2}{c}{\texttt{intervention\_layer}} \\
\cmidrule(l){2-3}
 & \textbf{Fold 1} & \textbf{Fold 2} \\
\midrule
OLMo2-7B-Instruct    & 19 & 11 \\
OLMo2-32B-Instruct   & 48 & 43 \\
Qwen2.5-7B-Instruct  & 17 & 19 \\
Qwen2.5-32B-Instruct & 47 & 40 \\
Llama3-8B-Instruct   & 9  & 18 \\
Llama3-70B-Instruct  & 39 & 39 \\
\bottomrule
\end{tabular}
\caption{Per-fold \texttt{intervention\_layer} selected for CAA and applied to the other.}
\label{tab:hp_caa}
\end{table}

\subsection{Steering Methods}
\label{sec:hp_steering}

\paragraph{ITI.}
For ITI, we follow the default setting of the original paper \cite{li2023inference}, using $K{=}48$ heads and an intervention strength of $\alpha{=}15$. As in the original work, we split TruthfulQA into training and validation sets to identify the top-$K$ heads, and then apply interventions on these heads at inference time on our benchmark.

\paragraph{CAA.} 
For CAA, we follow the original paper's procedure for selecting the intervention layer per model, using $\Delta p$, the change in the probability of the ground-truth answer's first token before and after intervention, as the selection criterion. Since aggregating over a larger number of contrast pairs reduces noise in the steering vector \cite{rimsky-etal-2024-steering}, we perform layer selection on all 3,254 facts prior to Corpus presence filtering. To prevent data leakage, we split this pool into two folds. Within each fold, 80\% is used to construct the steering vector and the remaining 20\% to select the optimal layer, which is then applied at inference on the other fold. As a result, every benchmark fact is evaluated with a steering vector and layer derived without it. We set the steering multiplier to $1.0$. The selected layers for each model are listed in Table~\ref{tab:hp_caa}.

\subsection{Prompting Methods}
\label{sec:recite_genread_prompts}

\paragraph{RECITE.}
Figure~\ref{fig:recite_stage1} and Figure~\ref{fig:recite_stage2} show the
prompts for stage 1 (recitation) and stage 2 (answer generation), respectively. Following the original paper's setting~\cite{sun2023recitationaugmented}, we run stage 1 and stage 2 for 20 times and choose the final answer with majority voting. 

\begin{figure}[t]
\centering
\begin{tcolorbox}[
  colback=gray!5, colframe=gray!50, fonttitle=\bfseries,
  title=RECITE Stage 1: Recitation Prompt,
  boxrule=0.5pt, left=4pt, right=4pt, top=4pt, bottom=4pt,
]
\small\ttfamily
Question: \{question\}\\[4pt]
The answer to the above question can be found in the following Wikipedia page, section, and paragraph:\\[4pt]
Answer: \{entity\} --- \{section\} --- Paragraph \#1\\
\{recited passage\}
\end{tcolorbox}
\caption{RECITE Stage 1 prompt. Five demonstrations of this form are provided in
the system message.}
\label{fig:recite_stage1}
\end{figure}
\begin{figure}[t]
\centering
\begin{tcolorbox}[
  colback=gray!5, colframe=gray!50, fonttitle=\bfseries,
  title=RECITE Stage 2: Answer Generation Prompt,
  boxrule=0.5pt, left=4pt, right=4pt, top=4pt, bottom=4pt,
]
\small\ttfamily
\{recited passage\}\\[4pt]
Based on the above paragraph, could you answer the following (probably) relevant questions?\\[4pt]
\{five few-shot demonstrations\}
\end{tcolorbox}
\caption{RECITE Stage 2 prompt.}
\label{fig:recite_stage2}
\end{figure}

\paragraph{GenRead.}
Figure~\ref{fig:genread_stage1} and Figure~\ref{fig:genread_stage2} show the
prompts for stage 1 (background generation) and stage 2 (answer generation),
respectively.

\begin{figure}[t]
\centering
\begin{tcolorbox}[
  colback=gray!5, colframe=gray!50, fonttitle=\bfseries,
  title=GenRead Stage 1: Background Generation Prompt,
  boxrule=0.5pt, left=4pt, right=4pt, top=4pt, bottom=4pt,
]
\small\ttfamily
Generate a background document from Wikipedia to answer the given question.\\[4pt]
\{question\}
\end{tcolorbox}
\caption{GenRead Stage 1 prompt (zero-shot).}
\label{fig:genread_stage1}
\end{figure}
\begin{figure}[t]
\centering
\begin{tcolorbox}[
  colback=gray!5, colframe=gray!50, fonttitle=\bfseries,
  title=GenRead Stage 2: Answer Generation Prompt,
  boxrule=0.5pt, left=4pt, right=4pt, top=4pt, bottom=4pt,
]
\small\ttfamily
Refer to the passage below and answer the following question with just one entity. \\[4pt]
Passage: \{generated passage\} \\[4pt]
\{five few-shot demonstrations\}
\end{tcolorbox}
\caption{GenRead Stage 2 prompt.}
\label{fig:genread_stage2}
\end{figure}

\section{Benchmark Prompt Details}
\label{sec:appC}

Figure~\ref{fig:gen_prompt} shows the generation prompt used for the baseline, and Table~\ref{tab:few_shots} shows the five few-shot demonstrations used for each short-answer question.

\begin{figure}[t]
\centering
\begin{tcolorbox}[
  colback=gray!5, colframe=gray!50, fonttitle=\bfseries,
  title=Generation Prompt Template,
  boxrule=0.5pt, left=4pt, right=4pt, top=4pt, bottom=4pt,
]
\small\ttfamily
Answer the following question with a short entity name. Do not explain.\\[6pt]
\{five few-shot demonstrations\}\\[6pt]
Question: \{question\}\\
Answer:
\end{tcolorbox}
\caption{Prompt template for generating the output.}
\label{fig:gen_prompt}
\end{figure}
\begin{table}[t]
\centering
\small
\begin{tabularx}{\columnwidth}{@{}X l@{}}
\toprule
\textbf{Question} & \textbf{Answer} \\
\midrule
What religious order was Koenigsbruck Abbey associated with? & Cistercian \\
\addlinespace
Which Olympic Games did John Garrels participate in? & 1908 Summer Olympics \\
\addlinespace
Which dynasty was Thutmose III part of? & eighteenth dynasty \\
\addlinespace
Who founded World Book and Copyright Day? & UNESCO \\
\addlinespace
Which organization is Llanelli Scarlets affiliated with? & Welsh Rugby Union \\
\bottomrule
\end{tabularx}
\caption{Five few-shot demonstrations used for the output generation.}
\label{tab:few_shots}
\end{table}



\section{Additional Results}
\label{sec:appD}

\begin{figure*}[t]
  \centering
  \includegraphics[width=\textwidth]{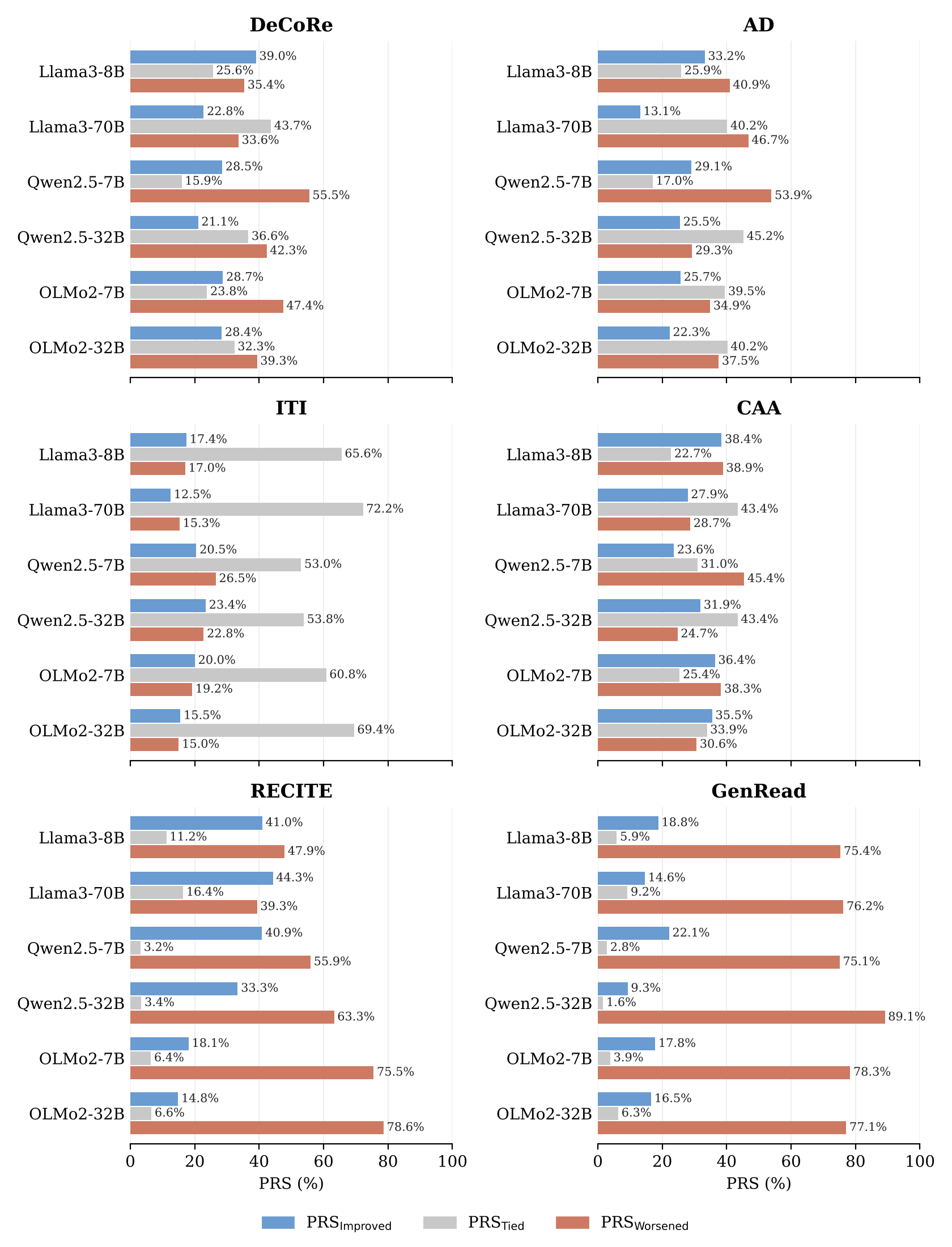}
  \caption{Per-model PRS on unrecovered failures in a bar graph.}
  \label{fig:prs_full}
\end{figure*}

Full results for raw count of Table~\ref{tab:tab1} are in Table~\ref{tab:raw_count}.
Full results for per-model and raw count breakdown of Figure~\ref{fig:prs} are in Figure~\ref{fig:prs_full} and Table~\ref{tab:prs_raw}.
Full results for raw count of Tables~\ref{tab:tab2} and~\ref{tab:sampling_t07} are in Tables~\ref{tab:depth_raw_count_07} and~\ref{tab:depth_raw_count_12}, respectively.

\begin{table*}[t]
\centering
\small
\renewcommand{\arraystretch}{1.2}
\begin{tabular}{ll cc cc cc cc | c}
\toprule
\multirow{2}{*}{\textbf{Model}} & \multirow{2}{*}{\textbf{Metric}} & \multicolumn{2}{c}{\textbf{Decoding}} & \multicolumn{2}{c}{\textbf{Steering}} & \multicolumn{2}{c}{\textbf{Prompting}} & \multicolumn{2}{c}{\textbf{Temperature Sampling}} & \multirow{2}{*}{\textbf{Denominator}} \\
\cmidrule(lr){3-4} \cmidrule(lr){5-6} \cmidrule(lr){7-8} \cmidrule(lr){9-10}
 & & \textbf{DeCoRe} & \textbf{AD} & \textbf{ITI} & \textbf{CAA} & \textbf{RECITE} & \textbf{GenRead} & \textbf{$\tau=0.7$} & \textbf{$\tau=1.2$} & \\
\midrule
\multirow{2}{*}{\textbf{OLMo2-7B}} & Gain$\uparrow$ & 23 & 13 & 34 & 21 & 161 & 138 & 23 & 25 & 692 \\
 & Loss$\downarrow$ & 23 & 23 & 21 & 31 & 39 & 63 & 17 & 44 & 611 \\
\midrule
\multirow{2}{*}{\textbf{OLMo2-32B}} & Gain$\uparrow$ & 23 & 15 & 14 & 28 & 129 & 112 & 15 & 33 & 479 \\
 & Loss$\downarrow$ & 19 & 27 & 12 & 35 & 49 & 67 & 28 & 36 & 824 \\
\midrule
\multirow{2}{*}{\textbf{Qwen2.5-7B}} & Gain$\uparrow$ & 32 & 28 & 9 & 11 & 153 & 185 & 10 & 26 & 741 \\
 & Loss$\downarrow$ & 56 & 63 & 23 & 17 & 51 & 69 & 9 & 28 & 562 \\
\midrule
\multirow{2}{*}{\textbf{Qwen2.5-32B}} & Gain$\uparrow$ & 8 & 6 & 9 & 12 & 126 & 131 & 8 & 19 & 572 \\
 & Loss$\downarrow$ & 13 & 4 & 16 & 8 & 55 & 84 & 8 & 14 & 731 \\
\midrule
 \multirow{2}{*}{\textbf{Llama3-8B}} & Gain$\uparrow$ & 72 & 65 & 30 & 71 & 126 & 186 & 19 & 40 & 644 \\
 & Loss$\downarrow$ & 36 & 45 & 25 & 59 & 49 & 81 & 15 & 44 & 659 \\
\midrule
\multirow{2}{*}{\textbf{Llama3-70B}} & Gain$\uparrow$ & 23 & 34 & 12 & 36 & 94 & 137 & 7 & 13 & 419 \\
 & Loss$\downarrow$ & 14 & 28 & 11 & 37 & 56 & 59 & 8 & 12 & 884 \\
\bottomrule
\end{tabular}
\caption{Raw counts for Table~\ref{tab:tab1}.}
\label{tab:raw_count}
\end{table*}
\begin{figure*}[t]
  \centering
  \begin{subfigure}[t]{0.49\textwidth}
    \centering
    \includegraphics[width=\linewidth]{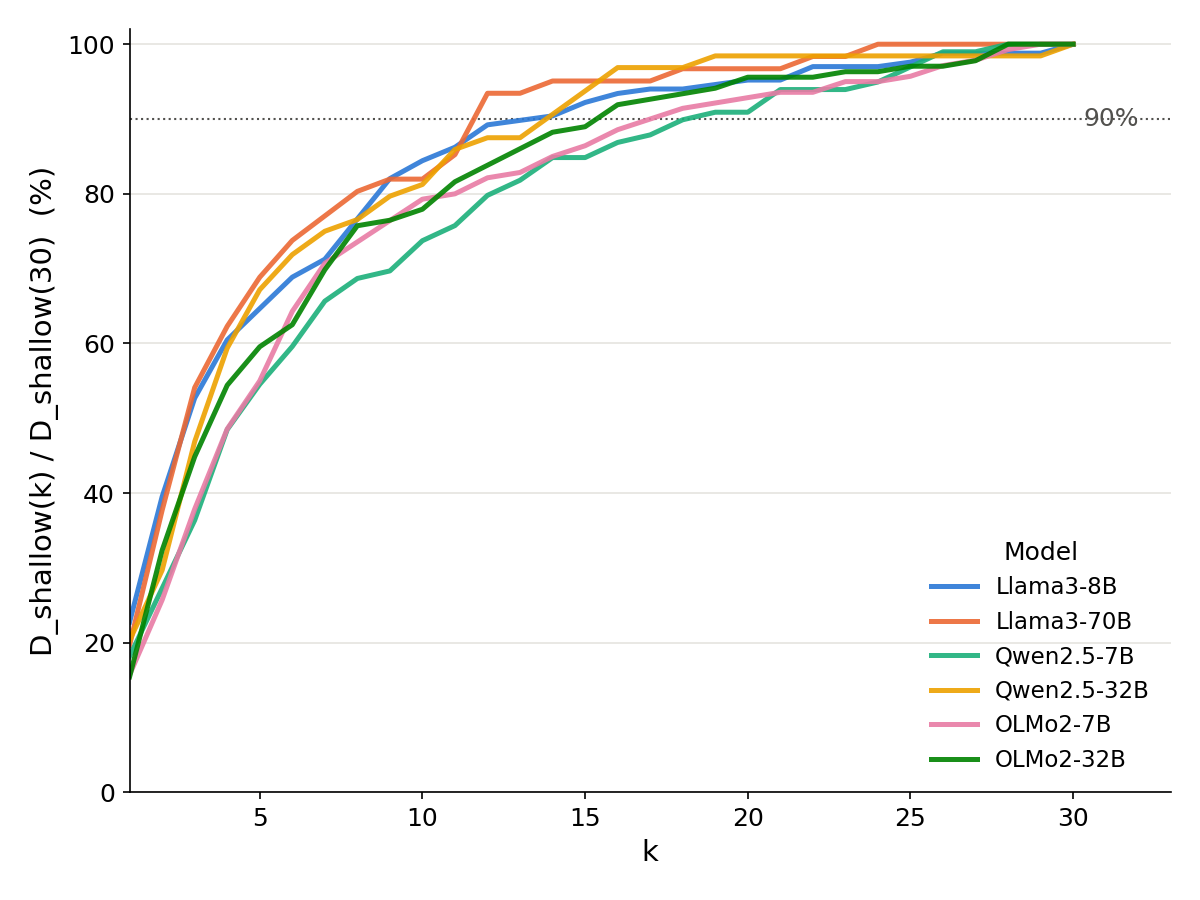}
    \caption{}
    \label{fig:k_t07}
  \end{subfigure}
  \hfill
  \begin{subfigure}[t]{0.49\textwidth}
    \centering
    \includegraphics[width=\linewidth]{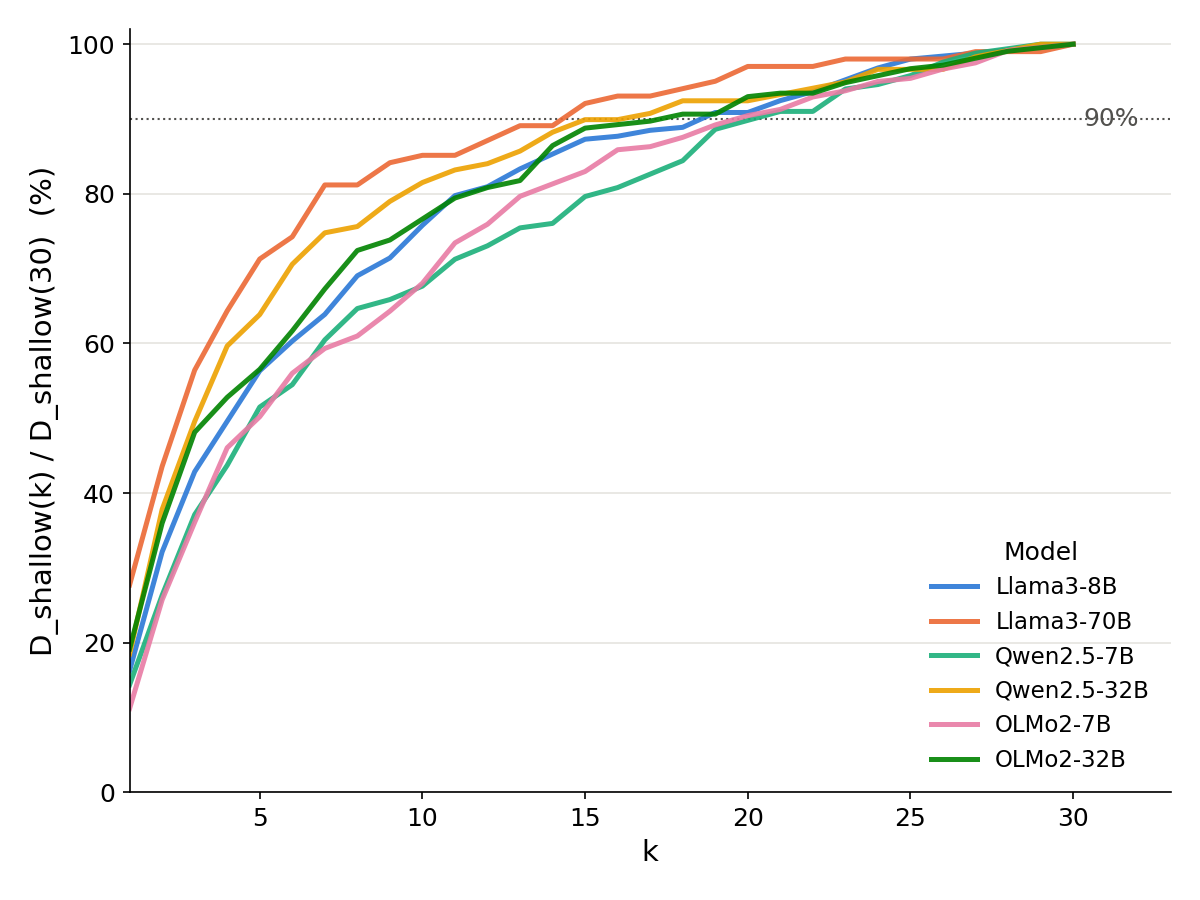}
    \caption{}
    \label{fig:k_t12}
  \end{subfigure}
\caption{Proportion of shallow failures identified with $K$ samples relative to those identified with 30 samples, $|\mathcal{D}_{\text{shallow}}(K)| / |\mathcal{D}_{\text{shallow}}(30)|$, for $K$ from 1 to 30. (a) The left panel shows temperature sampling at $\tau=0.7$ and (b) the right panel shows $\tau=1.2$. The dotted line marks 90\%. Coverage rises steeply for small $K$ and saturates afterward, and at $K=20$ every model reaches about 90\% or more at both temperatures. We therefore use $K=20$ as a balance between coverage and sampling cost.}
  \label{fig:k}
\end{figure*}

\begin{table*}[t]
\centering
\footnotesize
\renewcommand{\arraystretch}{1.0}
\setlength{\tabcolsep}{10pt}
\resizebox{\textwidth}{!}{%
\begin{tabular}{>{\centering\arraybackslash}m{1.8cm} | l r r | l r r | l r r}
\toprule
\textbf{Category} & \textbf{Model} & \textbf{SGR$\uparrow$} & \textbf{DGR$\uparrow$}
& \textbf{Model} & \textbf{SGR$\uparrow$} & \textbf{DGR$\uparrow$}
& \textbf{Model} & \textbf{SGR$\uparrow$} & \textbf{DGR$\uparrow$} \\
\midrule
\rowcolor{blue!8}
& \textbf{OLMo2-7B}       & --    & --    & \textbf{Qwen2.5-7B}    & --    & --    & \textbf{Llama3-8B}      & --    & --    \\
\textbf{Sampling}
& \quad + $\tau_{0.7}$ & 16.15 & 0.18  & \quad + $\tau_{0.7}$ & 11.11 & 0.00  & \quad + $\tau_{0.7}$ & 14.47 & 0.00  \\
\rowcolor{gray!8}
& \quad + DeCoRe & 16.15 & 0.36  & \quad + DeCoRe & 24.44 & 1.54  & \quad + DeCoRe & 32.70 & 4.12  \\
\rowcolor{gray!8}
\multirow{-2}{*}{\textbf{Decoding}}
& \quad + AD     & 10.00 & 0.00  & \quad + AD     & 23.33 & 1.08  & \quad + AD     & 27.67 & 4.33  \\
\multirow{2}{*}{\textbf{Steering}}
& \quad + ITI    & 17.69 & 1.96  & \quad + ITI    & 6.67  & 0.46  & \quad + ITI    & 12.58 & 2.06  \\
& \quad + CAA    & 15.38 & 0.18  & \quad + CAA    & 11.11 & 0.15  & \quad + CAA    & 29.56 & 4.95  \\
\rowcolor{gray!8}
& \quad + RECITE & \textbf{55.38} & \textbf{15.84} & \quad + RECITE & \underline{60.00} & \underline{15.21} & \quad + RECITE & \underline{47.80} & \underline{10.31} \\
\rowcolor{gray!8}
\multirow{-2}{*}{\textbf{Prompting}}
& \quad + GenRead& \underline{47.69} & \underline{13.52} & \quad + GenRead& \textbf{62.22} & \textbf{19.82} & \quad + GenRead& \textbf{60.38} & \textbf{18.56} \\
\addlinespace[3pt]\midrule\addlinespace[1pt]
\rowcolor{blue!8}
& \textbf{OLMo2-32B}      & --    & --    & \textbf{Qwen2.5-32B}   & --    & --    & \textbf{Llama3-70B}     & --    & --    \\
\textbf{Sampling}
& \quad + $\tau_{0.7}$ & 11.54 & 0.00  & \quad + $\tau_{0.7}$ & 15.87 & 0.00  & \quad + $\tau_{0.7}$ & 6.78  & 0.00  \\
\rowcolor{gray!8}
& \quad + DeCoRe & 16.15 & 0.57  & \quad + DeCoRe & 12.70 & 0.00  & \quad + DeCoRe & 25.42 & 2.22  \\
\rowcolor{gray!8}
\multirow{-2}{*}{\textbf{Decoding}}
& \quad + AD     & 11.54 & 0.00  & \quad + AD     & 9.52  & 0.00  & \quad + AD     & 30.51 & 4.44  \\
\multirow{2}{*}{\textbf{Steering}}
& \quad + ITI    & 10.77 & 0.00  & \quad + ITI    & 11.11 & 0.39  & \quad + ITI    & 16.95 & 0.56  \\
& \quad + CAA    & 20.77 & 0.29  & \quad + CAA    & 17.46 & 0.20  & \quad + CAA    & 35.59 & 4.17  \\
\rowcolor{gray!8}
& \quad + RECITE & \textbf{56.92} & \textbf{15.76} & \quad + RECITE & \textbf{55.56} & \underline{17.88} & \quad + RECITE & \underline{54.24} & \underline{17.22} \\
\rowcolor{gray!8}
\multirow{-2}{*}{\textbf{Prompting}}
& \quad + GenRead& \underline{53.08} & \underline{12.32} & \quad + GenRead& \underline{50.79} & \textbf{19.45} & \quad + GenRead& \textbf{62.71} & \textbf{27.78} \\
\bottomrule
\end{tabular}
}
\caption{SGR/DGR (\%) across model families and KEE techniques at $\tau{=}0.7$. \textbf{Bold} and \underline{underline} indicate the best and second-best values, respectively. Alternating gray shading indicates technique category boundaries.}
\label{tab:sampling_t07}
\end{table*}
\begin{table*}[t]
\centering
\small
\renewcommand{\arraystretch}{1.2}
\begin{tabular}{ll c cc cc cc |c}
\toprule
\multirow{2}{*}{\textbf{Model}} & \multirow{2}{*}{\textbf{Metric}} & \textbf{Temp.} & \multicolumn{2}{c}{\textbf{Decoding}} & \multicolumn{2}{c}{\textbf{Steering}} & \multicolumn{2}{c}{\textbf{Prompting}} & \multirow{2}{*}{\textbf{Denominator}} \\
\cmidrule(lr){4-5} \cmidrule(lr){6-7} \cmidrule(lr){8-9}
 & & \textbf{0.7} & \textbf{DeCoRe} & \textbf{AD} & \textbf{ITI} & \textbf{CAA} & \textbf{RECITE} & \textbf{GenRead} & \\
\midrule
\multirow{2}{*}{\textbf{OLMo2-7B}} & SGR$\uparrow$ & 21 & 21 & 13 & 23 & 20 & 72 & 62 & 130 \\
 & DGR$\uparrow$ & 1 & 2 & 0 & 11 & 1 & 89 & 76 & 562 \\
\midrule
\multirow{2}{*}{\textbf{OLMo2-32B}} & SGR$\uparrow$ & 15 & 21 & 15 & 14 & 27 & 74 & 69 & 130 \\
 & DGR$\uparrow$ & 0 & 2 & 0 & 0 & 1 & 55 & 43 & 349 \\
\midrule
\multirow{2}{*}{\textbf{Qwen2.5-7B}} & SGR$\uparrow$ & 10 & 22 & 21 & 6 & 10 & 54 & 56 & 90 \\
 & DGR$\uparrow$ & 0 & 10 & 7 & 3 & 1 & 99 & 129 & 651 \\
\midrule
\multirow{2}{*}{\textbf{Qwen2.5-32B}} & SGR$\uparrow$ & 10 & 8 & 6 & 7 & 11 & 35 & 32 & 63 \\
 & DGR$\uparrow$ & 0 & 0 & 0 & 2 & 1 & 91 & 99 & 509 \\
\midrule
\multirow{2}{*}{\textbf{Llama3-8B}} & SGR$\uparrow$ & 23 & 52 & 44 & 20 & 47 & 76 & 96 & 159 \\
 & DGR$\uparrow$ & 0 & 20 & 21 & 10 & 24 & 50 & 90 & 485 \\
\midrule
\multirow{2}{*}{\textbf{Llama3-70B}} & SGR$\uparrow$ & 4 & 15 & 18 & 10 & 21 & 32 & 37 & 59 \\
 & DGR$\uparrow$ & 0 & 8 & 16 & 2 & 15 & 62 & 100 & 360 \\
\bottomrule
\end{tabular}
\caption{Raw counts for Table~\ref{tab:sampling_t07}.}
\label{tab:depth_raw_count_07}
\end{table*}
\begin{table*}[t]
\centering
\small
\renewcommand{\arraystretch}{1.2}
\begin{tabular}{ll c cc cc cc |c}
\toprule
\multirow{2}{*}{\textbf{Model}} & \multirow{2}{*}{\textbf{Metric}} & \textbf{Temp.} & \multicolumn{2}{c}{\textbf{Decoding}} & \multicolumn{2}{c}{\textbf{Steering}} & \multicolumn{2}{c}{\textbf{Prompting}} & \multirow{2}{*}{\textbf{Denominator}} \\
\cmidrule(lr){4-5} \cmidrule(lr){6-7} \cmidrule(lr){8-9}
 & & \textbf{1.2} & \textbf{DeCoRe} & \textbf{AD} & \textbf{ITI} & \textbf{CAA} & \textbf{RECITE} & \textbf{GenRead} & \\
\midrule
\multirow{2}{*}{\textbf{OLMo2-7B}} & SGR$\uparrow$ & 35 & 21 & 13 & 32 & 21 & 113 & 97 & 218 \\
 & DGR$\uparrow$ & 2 & 2 & 0 & 2 & 0 & 48 & 41 & 474 \\
\midrule
\multirow{2}{*}{\textbf{OLMo2-32B}} & SGR$\uparrow$ & 26 & 21 & 14 & 14 & 28 & 97 & 86 & 199 \\
 & DGR$\uparrow$ & 0 & 2 & 1 & 0 & 0 & 32 & 26 & 280 \\
\midrule
\multirow{2}{*}{\textbf{Qwen2.5-7B}} & SGR$\uparrow$ & 28 & 24 & 23 & 9 & 10 & 80 & 90 & 150 \\
 & DGR$\uparrow$ & 1 & 8 & 5 & 0 & 1 & 73 & 95 & 591 \\
\midrule
\multirow{2}{*}{\textbf{Qwen2.5-32B}} & SGR$\uparrow$ & 16 & 8 & 6 & 9 & 12 & 57 & 46 & 110 \\
 & DGR$\uparrow$ & 1 & 0 & 0 & 0 & 0 & 69 & 85 & 462 \\
\midrule
\multirow{2}{*}{\textbf{Llama3-8B}} & SGR$\uparrow$ & 36 & 55 & 48 & 23 & 54 & 95 & 121 & 229 \\
 & DGR$\uparrow$ & 1 & 17 & 17 & 7 & 17 & 31 & 65 & 415 \\
\midrule
\multirow{2}{*}{\textbf{Llama3-70B}} & SGR$\uparrow$ & 16 & 18 & 26 & 11 & 29 & 46 & 58 & 98 \\
 & DGR$\uparrow$ & 0 & 5 & 8 & 1 & 7 & 48 & 79 & 321 \\
\bottomrule
\end{tabular}
\caption{Raw counts for Table~\ref{tab:tab2}.}
\label{tab:depth_raw_count_12}
\end{table*}
\begin{table*}[t]
\centering
\small
\renewcommand{\arraystretch}{1.15}
\setlength{\tabcolsep}{5pt}
\begin{tabular}{ll cc cc cc}
\toprule
\multirow{2}{*}{\textbf{Model}} & \multirow{2}{*}{\textbf{Category}} & \multicolumn{2}{c}{\textbf{Decoding}} & \multicolumn{2}{c}{\textbf{Steering}} & \multicolumn{2}{c}{\textbf{Prompting}} \\
\cmidrule(lr){3-4} \cmidrule(lr){5-6} \cmidrule(lr){7-8}
 & & \textbf{DeCoRe} & \textbf{AD} & \textbf{ITI} & \textbf{CAA} & \textbf{RECITE} & \textbf{GenRead} \\
\midrule
\multirow{6}{*}{\textbf{OLMo2-7B}} & $n_{00}$ & 669 & 679 & 658 & 671 & 531 & 554 \\
 & $n_{\text{exc}}$ & 102 & 91 & 74 & 88 & 61 & 88 \\
 & $n_{\text{inc}}$ & 567 & 588 & 584 & 583 & 470 & 466 \\
 & Improved & 28.7 (163) & 25.7 (151) & 20.0 (117) & 36.4 (212) & 18.1 (85)  & 17.8 (83) \\
 & Tied     & 23.8 (135) & 39.5 (232) & 60.8 (355) & 25.4 (148) & 6.4 (30)   & 3.9 (18) \\
 & Worsened & 47.4 (269) & 34.9 (205) & 19.2 (112) & 38.3 (223) & 75.5 (355) & 78.3 (365) \\
\midrule
\multirow{6}{*}{\textbf{OLMo2-32B}} & $n_{00}$ & 456 & 464 & 465 & 451 & 350 & 367 \\
 & $n_{\text{exc}}$ & 100 & 96 & 79 & 82 & 60 & 83 \\
 & $n_{\text{inc}}$ & 356 & 368 & 386 & 369 & 290 & 284 \\
 & Improved & 28.4 (101) & 22.3 (82)  & 15.5 (60)  & 35.5 (131) & 14.8 (43)  & 16.5 (47) \\
 & Tied     & 32.3 (115) & 40.2 (148) & 69.4 (268) & 33.9 (125) & 6.6 (19)   & 6.3 (18) \\
 & Worsened & 39.3 (140) & 37.5 (138) & 15.0 (58)  & 30.6 (113) & 78.6 (228) & 77.1 (219) \\
\midrule
\multirow{6}{*}{\textbf{Qwen2.5-7B}} & $n_{00}$ & 709 & 713 & 732 & 730 & 588 & 556 \\
 & $n_{\text{exc}}$ & 113 & 108 & 102 & 107 & 209 & 95 \\
 & $n_{\text{inc}}$ & 596 & 605 & 630 & 623 & 379 & 461 \\
 & Improved & 28.5 (170) & 29.1 (176) & 20.5 (129) & 23.6 (147) & 40.9 (155) & 22.1 (102) \\
 & Tied     & 15.9 (95)  & 17.0 (103) & 53.0 (334) & 31.0 (193) & 3.2 (12)   & 2.8 (13) \\
 & Worsened & 55.5 (331) & 53.9 (326) & 26.5 (167) & 45.4 (283) & 55.9 (212) & 75.1 (346) \\
\midrule
\multirow{6}{*}{\textbf{Qwen2.5-32B}} & $n_{00}$ & 564 & 566 & 563 & 560 & 446 & 441 \\
 & $n_{\text{exc}}$ & 89 & 84 & 80 & 83 & 182 & 74 \\
 & $n_{\text{inc}}$ & 475 & 482 & 483 & 477 & 264 & 367 \\
 & Improved & 21.1 (100) & 25.5 (123) & 23.4 (113) & 31.9 (152) & 33.3 (88)  & 9.3 (34) \\
 & Tied     & 36.6 (174) & 45.2 (218) & 53.8 (260) & 43.4 (207) & 3.4 (9)    & 1.6 (6) \\
 & Worsened & 42.3 (201) & 29.3 (141) & 22.8 (110) & 24.7 (118) & 63.3 (167) & 89.1 (327) \\
\midrule
\multirow{6}{*}{\textbf{Llama3-8B}} & $n_{00}$ & 572 & 579 & 614 & 573 & 518 & 458 \\
 & $n_{\text{exc}}$ & 131 & 127 & 103 & 133 & 169 & 101 \\
 & $n_{\text{inc}}$ & 441 & 452 & 511 & 440 & 349 & 357 \\
 & Improved & 39.0 (172) & 33.2 (150) & 17.4 (89)  & 38.4 (169) & 41.0 (143) & 18.8 (67) \\
 & Tied     & 25.6 (113) & 25.9 (117) & 65.6 (335) & 22.7 (100) & 11.2 (39)  & 5.9 (21) \\
 & Worsened & 35.4 (156) & 40.9 (185) & 17.0 (87)  & 38.9 (171) & 47.9 (167) & 75.4 (269) \\
\midrule
\multirow{6}{*}{\textbf{Llama3-70B}} & $n_{00}$ & 396 & 385 & 407 & 383 & 325 & 282 \\
 & $n_{\text{exc}}$ & 128 & 126 & 119 & 125 & 203 & 97 \\
 & $n_{\text{inc}}$ & 268 & 259 & 288 & 258 & 122 & 185 \\
 & Improved & 22.8 (61)  & 13.1 (34)  & 12.5 (36)  & 27.9 (72)  & 44.3 (54)  & 14.6 (27) \\
 & Tied     & 43.7 (117) & 40.2 (104) & 72.2 (208) & 43.4 (112) & 16.4 (20)  & 9.2 (17) \\
 & Worsened & 33.6 (90)  & 46.7 (121) & 15.3 (44)  & 28.7 (74)  & 39.3 (48)  & 76.2 (141) \\
\bottomrule
\end{tabular}
\caption{Per-model PRS on unrecovered failures. Among the $n_{00}$ unrecovered failures, $n_{\text{exc}}$ are excluded because the gold token is ranked first by either the baseline or the KEE technique, and for RECITE this includes ranking first in any of its 20 recitation paths. The remaining $n_{\text{inc}} = N_{\text{Improved}} + N_{\text{Tied}} + N_{\text{Worsened}}$ failures are categorized by Eq.~\ref{eq:prs}, and each category reports $\text{PRS}_c$ in percent followed by $N_c$ in parentheses. For RECITE, each failure is assigned the most frequent category across its 20 recitation paths.}
\label{tab:prs_raw}
\end{table*}

\section{Use of AI assistants}

AI assistants like ChatGPT and Claude Code were intermittently used to generate template code and rephrase sentences in the paper, etc. However, no complete paper sections or code scripts have been generated by an AI assistant. All generated content has been inspected and verified by the authors.
\end{document}